\documentclass[sigconf, nonacm]{acmart}
\usepackage{stfloats}
\usepackage{appendix}
\usepackage{multirow}
\usepackage{subfig}
\usepackage{color}
\usepackage{booktabs} % For formal tables
\usepackage{enumitem}
\usepackage{balance}
\usepackage{makecell}
\usepackage{threeparttable}
\usepackage{amsmath,amsfonts,mathtools}%,amssymb}
\usepackage{mathrsfs}
\usepackage{float}
\usepackage{enumitem}
\usepackage{graphicx}%#\usepackage{algorithm}

\usepackage[linesnumbered,vlined,ruled]{algorithm2e}
\usepackage{enumitem}
\usepackage{graphicx}
\usepackage{ulem}
\usepackage{tabularx}
\usepackage{lineno}

\begin{document}
\title{LOGIN: A Large Language Model Consulted Graph Neural Network Training Framework}

%%
%% The "author" command and its associated commands are used to define the authors and their affiliations.
\author{Yiran Qiao}
\authornotemark[2]
\affiliation{%
  \institution{Institute of Computing Technology,
Chinese Academy of Sciences\\ University of Chinese Academy of Sciences}
  \city{Beijing}
  \country{China}
  \postcode{100190}
}
\email{yrqiao@gmail.com}

\author{Xiang Ao}
\authornote{Corresponding author.}
\authornotemark[2]
\affiliation{%
  \institution{Institute of Computing Technology,
Chinese Academy of Sciences\\ University of Chinese Academy of Sciences}
  \city{Beijing}
  \country{China}}
\email{aoxiang@ict.ac.cn}

\author{Yang Liu}
\authornotemark[2]
\affiliation{%
  \institution{Institute of Computing Technology,
Chinese Academy of Sciences\\ University of Chinese Academy of Sciences}
  \city{Beijing}
  \country{China}
  \postcode{100190}
}
\email{liuyang2023@ict.ac.cn}

\author{Jiarong Xu}
\affiliation{%
  \institution{School of Management, Fudan University}
  \city{Shanghai}
  \country{China}
}
\email{jiarongxu@fudan.edu.cn}

\author{Xiaoqian Sun}
\authornotemark[2]
\affiliation{%
  \institution{Institute of Computing Technology,
Chinese Academy of Sciences\\ University of Chinese Academy of Sciences}
  \city{Beijing}
  \country{China}
  \postcode{100190}
}
\email{sunxiaoqian@ict.ac.cn}

\author{Qing He}
\authornote{Key Lab of AI Safety of Chinese Academy of Sciences (CAS). Xiang Ao is also at Institute of Intelligent Computing Technology, Suzhou, China.}
\affiliation{%
  \institution{Institute of Computing Technology,
Chinese Academy of Sciences\\ University of Chinese Academy of Sciences}
  \city{Beijing}
  \country{China}
  \postcode{100190}
}
\email{heqing@ict.ac.cn}

\begin{abstract}
  Recent prevailing works on graph machine learning typically follow a similar methodology that involves designing advanced variants of graph neural networks~(GNNs) to maintain the superior performance of GNNs on different graphs. In this paper, we aim to streamline the GNN design process and leverage the advantages of Large Language Models~(LLMs) to improve the performance of GNNs on downstream tasks. 
  We formulate a new paradigm, coined ``LLMs-as-Consultants'', which integrates LLMs with GNNs in an interactive manner. A framework named \textbf{L\scalebox{1}[0.8]{O}G\scalebox{1}[0.8]{IN}}~(\underline{\textbf{L}}\scalebox{1}[0.8]{LM} \scalebox{1}[0.9]{c}\underline{\textbf{O}}\scalebox{1}[0.9]{nsulted} \underline{\textbf{G}}\scalebox{1}[0.8]{NN} \scalebox{1}[0.9]{tra}\underline{\textbf{IN}}\scalebox{1}[0.9]{ing}) is instantiated, empowering the interactive utilization of LLMs within the GNN training process. 
  First, we attentively craft concise prompts for spotted nodes, carrying comprehensive semantic and topological information, and serving as input to LLMs. Second, we refine GNNs by devising a complementary coping mechanism that utilizes the responses from LLMs, depending on their correctness. We empirically evaluate the effectiveness of L\scalebox{1}[0.8]{O}G\scalebox{1}[0.8]{IN} on node classification tasks across both homophilic and heterophilic graphs. The results illustrate that even basic GNN architectures, when employed within the proposed LLMs-as-Consultants paradigm, can achieve comparable performance to advanced GNNs with intricate designs. Our codes are available at \url{https://github.com/QiaoYRan/LOGIN}.
\end{abstract}

\maketitle

\section{Introduction}\label{sec:intro}
Individual entities along with their respective interactions, typically denoting as $\langle{u}, e_{u,v},{v}\rangle$, where $u$ and $v$ are entities and $e_{u,v}$ is the interaction between $u$ and $v$, can be effectively represented and analyzed in the form of graph-structured data. 
For instance, reciprocal following in social networks~\cite{2006Inferring,2007Topic}, financial transfers between accounts~\cite{akoglu2014graphbased,pourhabibi2020fraud}, chemical bonds between molecules~\cite{blundell1996structure,kearnes2016molecular}, and citations of academic articles~\cite{leskovec2005graphs,alcacer2006patent} can be generally represented as graph-like structures.  
It thus burgeons graph machine learning in various tasks for decades, such as recommendation~\cite{ying2018graph,ma2019learning,1998Recommendation}, anomaly detection~\cite{liu2021pick,zhong2020financial,Christopher2002Service}, drug discovery~\cite{stark20223d,wang2023scientific,lo2018machine}, and etc~\cite{hamilton2017representation,riesen2008iam,Broder2000Graph}.
Recently, thanks to the advent of graph neural networks~(GNNs)~\cite{kipf2016semi,hamilton2017inductive,velickovic2017graph} and their demonstrated extensive capabilities, graph machine learning has become a highly active field attracting significant research attention. 
%\redfont{More than one-third of accepted papers in the research track of KDD’23 explicitly contain the term  ``graph'' in their titles, underscoring the significant popularity of this topic}\footnote{\url{https://kdd.org/kdd2023/research-track-papers/}}.
%Owing to this extensive applicability, the field of graph machine learning is highly active with significant research attention. As in the papers accepted for the research track at KDD 2023, more than one-third of the titles explicitly feature the term "graph". 

Among these existing works, most of them strive to design different architectures of GNN to adapt to distinct graph types. The reason for that is the broad real-world sources of graph data induce considerable diversity across different graphs. One of the basic categorizations, for example, could be the case of homophily and heterophily~\cite{1970Homophily,rogers1970homophily}. 
Classic GNNs, due to their fundamental mechanisms of message passing and aggregation~\cite{battaglia2018relational}, have demonstrated superior performance on homophilic graphs but fail to generalize to heterophily scenarios where dissimilar nodes are connected~\cite{zhu2020beyond,luan2020complete,liu2022ud}. Consequently, researchers have committed to developing various structures, aiming at improving the effectiveness of GNNs on particular types of graphs~\cite{lim2021large,liu2022ud,pei2019geom,luan2020complete}.
%traditional GNNs failed on heterophilic graphs where dissimilar nodes bond~\cite{zhu2020beyond,luan2020complete}.
In addition to individualized designs, there are also observed recent works applying Graph Neural Architecture Search~(GNAS) techniques for seeking better GNN architecture from a series of essential components~\cite{zhou2022auto,zhang2023unsupervised}. 
%\redfont{\reminder{move conclusion backwards.}}
Despite their various routes of the existing works, these methods hold a similar methodology: Design a specialized variant of graph neural networks to tackle specific analysis tasks.

%For instance, in the case of homophily and heterophily, such a categorization within graphs are not artificially constructed. Instead, they reflect the inherent properties observed in real-world scenarios~\cite{pandit2007netprobe,zhu2020beyond,zhu2021graph}. This variation in graph structures has a profound impact on the performance of graph models. Traditional Graph Neural Networks~(GNNs) have demonstrated superior performance compared to Neural Networks~(NNs) due to their fundamental mechanisms of message passing and aggregation~\cite{battaglia2018relational}. However, when the homophily assumption does not hold, traditional GNNs failed on heterophilic graphs where dissimilar nodes bond~\cite{zhu2020beyond,luan2020complete}. Consequently, researchers have committed to developing a variety of specialized structures, aiming at improving the effectiveness of GNNs on particular types of graphs~\cite{lim2021large,liu2022ud,pei2019geom,luan2020complete}. Graph Neural Architecture Search~(GNAS) is also applied to determine the best GNN architecture for different datasets~\cite{zhou2022auto,zhang2023unsupervised}. Despite their success, these methods present huge complexity both theoretically and technically. It is challenging to distinguish which parts of the model are pivotal to its power. So the question naturally arises: \emph{Adherent to Occam's Razor, could we find an existing tool that empowers GNNs to learn over various datasets in a simple yet effective way?}

%%(Large language models~(LLMs) have shown their distinctive capabilities in various natural language tasks.) 

However, designing these GNN variants to work effectively in real-world applications requires extensive manual experimentation. Researchers and practitioners may engage in a meticulous process of trial and error when adjusting model architectures. This process is usually time-consuming and demands a deep understanding of the theoretical underpinnings of GNNs and the practical requirements of business. Achieving a model that works and performs efficiently and accurately in specific scenarios often involves long periods of rigorous experimentation and iterative refinement. Moreover, these GNN variants tailored for different scenarios are not significantly different from the classic GNN models theoretically~\cite{zhu2021interpreting}. In other words, though they may incorporate modifications to address specific needs or optimize performance for particular graphs, the foundational principles and underlying architectures might remain largely consistent with classic GNNs. Whether the capacities of classic GNNs can be maximized for different scenarios has become an open research problem.

%In this paper, \redfont{our objective is to streamline the GNN design process and answer the question of whether diverse graph-structured data can be addressed using a simple and unified architecture, e.g. classic GNNs.} 
In this paper, our objective is to empower classic GNNs for consistently superior performance on graphs with varying characteristics.
To this end, an influential off-shelf tool, the Large Lanuange Model~(LLM), is adopted to help us achieve the goal. 
Following the scaling law~\cite{kaplan2020scaling}, LLMs, such as 175B-parameter GPT-3~\cite{brown2020language}, display distinct behaviors compared to smaller pre-trained language models~(PLMs) like the 330M-parameter BERT~\cite{devlin2018bert} and the 1.5B-parameter GPT-2~\cite{radford2019language}. LLMs exhibit surprising emergent abilities~\cite{wei2022emergent}, enabling them to solve complex tasks that smaller models struggle with~\cite{zhao2023survey}. Consequently, the research community has focused increasingly on these LLMs, acknowledging their unique and advanced capabilities~\cite{brown2020language,touvron2023llama,alayrac2022flamingo,liu2023visual,ren2023representation,mysore2023large}.

LLMs currently have not only achieved remarkable advancements in natural language processing~\cite{brown2020language,touvron2023llama}, but have also buoyed promising insights in other domains, such as computer vision~\cite{alayrac2022flamingo,liu2023visual,liu2024sora,reid2024gemini} and recommendation systems~\cite{ren2023representation,mysore2023large,lin2024clickprompt,zhu2024collaborative}, etc. This success can be attributed to the wealth of open-world knowledge stored in their large-scale parameters~\cite{zhao2023survey,pan2024unifying} and the emerging capabilities in logical reasoning~\cite{huang2022towards,brown2020language}. 
Analogously, we aim to leverage the advantages of LLMs to enhance graph machine learning, particularly the capacities of GNNs. 
We formulate a new paradigm that integrates LLMs with GNNs in an interactive manner, and we coin it ``LLMs-as-Consultants'', following which, LLMs could explicitly contribute to the training process of GNN.

%and relying solely on LLMs to classify yields only moderate performance, disregarding the advantages of graph machine learning models, e.g. GNN. For LLMs-as-Enhancers, through distillation and fine-tuning, smaller language models gain insights from LLMs to enhance node features semantically~\cite{chen2023exploring,he2023harnessing,yu2023empower}. Utilizing LLMs to assess the soundness of links for local structure enhancement is also investigated~\cite{sun2023large}. 

Under this paradigm, we propose a framework named \textbf{L\scalebox{1}[0.8]{O}G\scalebox{1}[0.8]{IN}}, short for \underline{\textbf{L}}\scalebox{1}[0.8]{LM} \scalebox{1}[0.9]{c}\underline{\textbf{O}}\scalebox{1}[0.9]{nsulted} \underline{\textbf{G}}\scalebox{1}[0.8]{NN} \scalebox{1}[0.9]{tra}\underline{\textbf{IN}}\scalebox{1}[0.9]{ing}, empowering the interactive utilization of LLMs within the GNN training process. As an interactive approach, the crucial issues of L\scalebox{1}[0.8]{O}G\scalebox{1}[0.8]{IN} lie in what GNNs should deliver to LLMs and how to feed the LLMs' responses back to GNNs. First, for LLMs' inputs, we delve into prompt engineering to craft concise prompts for spotted nodes, which carry comprehensive semantic and topological information. Second, to utilize the responses from LLMs, we devise a complementary coping mechanism depending on their correctness. Specifically, compared to ground truth labels, when LLMs predict correctly, we update node features to obtain semantic enhancement. Otherwise, we impute the misclassification to the potential presence of local topological noises, hence performing structure refinement. Besides, particular criteria may serve in the selection of essential nodes to reduce time complexity. In our implementation, we adopt GNN predictive uncertainty to assess the necessity of consulting LLMs. 

To demonstrate the versatility and applicability of the proposed framework, we explore the effectiveness of L\scalebox{1}[0.8]{O}G\scalebox{1}[0.8]{IN} on node classification tasks across both homophilic and heterophilic graphs. By empirical studies, we illustrate that even classic GNNs, when employed within the proposed LLMs-as-Consultants paradigm, can achieve comparable performance to advanced GNNs with intricate designs.

The contributions of this paper can be summarized as follows.
\begin{itemize}
    \item To our knowledge, we are the first to propose the LLMs-as-Consultants paradigm of graph machine learning. Different from previous works, we integrate the power of LLMs interactively into GNN training.
    \item Under this paradigm, we propose the \underline{\textbf{L}}\scalebox{1}[0.8]{LM} \scalebox{1}[0.9]{c}\underline{\textbf{O}}\scalebox{1}[0.9]{nsulted} \underline{\textbf{G}}\scalebox{1}[0.8]{NN} \scalebox{1}[0.9]{tra}\underline{\textbf{IN}}\scalebox{1}[0.9]{ing}~(\textbf{L\scalebox{1}[0.8]{O}G\scalebox{1}[0.8]{IN}}) framework. This framework can be considered as a synthesis of previous methodologies, with a particularly tailored feedback strategy concerning the correctness of responses.
    \item Experiments on six node classification tasks with both homophilic and heterophilic graphs demonstrate the effectiveness and generalizability of L\scalebox{1}[0.8]{O}G\scalebox{1}[0.8]{IN}.
\end{itemize}
The remainder of this paper is organized as follows. Section~\ref{sec:relatedwork} surveys the related research in literature. Section~\ref{sec:discussion} discusses the core differences and advantages of our proposed LLMs-as-Consultants paradigm, distinguishing it from previous methods.
Section~\ref{sec:preliminary} introduces the concepts and preliminaries of this work. Section~\ref{sec:method} details our L\scalebox{1}[0.8]{O}G\scalebox{1}[0.8]{IN} approach. Section~\ref{sec:exp} illustrates the evaluation of our model and baselines in six benchmark datasets, and Section~\ref{sec:conclusion} concludes the paper and discusses future work. 
\section{Related Works}\label{sec:relatedwork}
\subsection{GNN Variants}
%%Early in 1970, the concept of heterophily~\cite{rogers1970homophily} was first introduced for communication research, which refers to the difference between the characteristics of communicators and corresponding recipients. Abstracted into the graph field, this concept implies that dissimilar nodes are likely to establish links. For graph-structured data, the classic GNNs outperform ordinary neural networks due to the message passing mechanism, but only when the homophily assumption holds.
%Introduced early in 2019, MixHop~\cite{abu2019mixhop} stands as an early exemplary approach for adapting GNNs to heterophily, by broadening the aggregation mechanism to encompass multi-hop neighborhoods. This innovation significantly boosts the advancement of heterophilic graph learning. Generally, the specialized designs for tailoring GNNs to heterophily can be categorized into neighbor extension and GNN architecture refinement~\cite{zheng2022graph}. 
Recall that early proposed GNNs~\cite{kipf2016semi,hamilton2017inductive,velickovic2017graph} are primarily designed for the most typical graph-structured data, such as citation networks, which are characterized as homophilic, homogeneous, and class-balanced graphs. Here homophily means nodes with similar attributes are more likely to connect~\cite{hamilton2020graph,1970Homophily}. For the second one, homogeneous graphs compose different types of entities (i.e.,~nodes) and relations (i.e.,~edges)~\cite{wang2019heterogeneous,zhang2019heterogeneous}. Class balance refers to the situation where the distribution of classes or labels within a graph is nearly even~\cite{zhao2021graphsmote,ju2024survey}. Consequently, when generalized to heterophilic~\cite{bo2021beyond,abu2019mixhop,xu2018representation,chen2020simple}, heterogeneous~\cite{zhang2019heterogeneous,wang2019heterogeneous,yun2019graph,fu2020magnn},  and class-imbalanced graphs~\cite{zhao2021graphsmote,qu2021imgagn,park2021graphens,liu2021pick,chen2021topology,song2022tam}, classic GNNs suffer from severe performance degradation. 

However, the characteristics of heterophily, heterogeneity, and class imbalance are widely recognized in the graphs constructed from real-world scenarios.
For example, heterophily is evident in online transaction networks as fraudsters are more likely to form connections with normal customers~\cite{huang2022auc}; in molecular graphs, different
types of amino acids tend to link as to compose proteins~\cite{zhu2020beyond}. Heterogeneous graphs can represent the various entities and their relations in recommendation~\cite{shi2018heterogeneous,hu2018leveraging} and cybersecurity systems~\cite{hu2019cash,hou2017hindroid}. Class imbalance is present in fraud detection based on financial
transaction graphs~\cite{liu2021pick,tian2023asa} due to the rarity of fraudulent cases.

%In 2019, MixHop~\cite{abu2019mixhop} was introduced as an early exemplary approach for adapting GNNs to heterophily, broadening the aggregation mechanism to encompass multi-hop neighborhoods. This innovation significantly advances heterophilic graph learning. 

To address the challenges GNNs face in generalizing to a wide range of real-world applications, researchers have been motivated to tailor classic GNNs to specific scenarios, namely heterophily, heterogeneity, class imbalance, and so on.
 
 \subsubsection{GNNs Designed for Heterophily} For heterophily, since similar nodes may not tend to bond, neighbor extension and GNN architecture refinement are two major designs for variants targeting heterophilic graphs. For neighbor extension, MixHop~\cite{abu2019mixhop} and H2GCN~\cite{zhu2020beyond} enable GNNs to aggregate information across multi-hop ego-graphs. 
 For GNN architecture refinement, H2GCN~\cite{zhu2020beyond} also excludes self-loop connections for non-mixing node embedding learning to produce distinguishable node embeddings. JKNet~\cite{xu2018representation} combines intermediate representations from each layer, allowing flexible adaptation of neighborhood ranges to individual nodes. H2GCN~\cite{zhu2020beyond} and GCNII~\cite{chen2020simple} further utilize all intermediate representations and only the first layer’s node embedding at each layer with the initial residual connection to realize inter-layer combination, respectively.

  \subsubsection{GNNs Designed for Heterogeneity}
 For heterogeneity, the key to designing a specified GNN variant is to propose a suitable aggregation
function~\cite{wang2022survey}, which can better fuse the structural and semantic information of heterogeneous graphs. To name some, HAN~\cite{wang2019heterogeneous} develops a hierarchical attention mechanism to determine the significance of various nodes and meta-paths, effectively capturing both the structural and semantic information within heterogeneous graphs. HetGNN~\cite{zhang2019heterogeneous} employs bi-LSTM to aggregate neighbor embeddings, facilitating the learning of deep interactions among heterogeneous nodes. 
GTN~\cite{yun2019graph} proposes an aggregation function that can automatically discover suitable metapaths during the message-passing process. MAGNN~\cite{fu2020magnn} designs three main components, i.e.,~node content transformation for input node attributes, intra-metapath aggregation for intermediate semantic nodes, and inter-metapath aggregation to integrate messages from multiple metapaths, with the aim to enhance its expressing power of heterogeneous graphs.

 \subsubsection{GNNs Designed for Class Imbalance} 
For class imbalance, the goal of designing GNN variants is to develop a GNN classifier that effectively handles both majority and minority classes~\cite{ju2024survey}. 
For example, GraphSMOTE~\cite{zhao2021graphsmote} utilizes synthetic minority oversampling in the embedding space to boost minority class representation. ImGAGN~\cite{qu2021imgagn} and GraphENS~\cite{park2021graphens} address the class imbalance by generating synthetic minority nodes. PC-GNN~\cite{liu2021pick} utilizes a label-balanced sampler to pick nodes for sub-graph training and a neighborhood sampler to balance the local topology.
ReNode~\cite{chen2021topology} and TAM~\cite{song2022tam} adjust the weights of labeled nodes in the loss function utilizing topological information.

Theoretically, the aforementioned GNN variants tailored for different scenarios are not significantly different from the classic GNNs~\cite{zhu2021interpreting}. Practically, designing these variants to work is also challenging; it requires extensive manual experimentation over a long period. Our work aims to leverage LLMs to optimize GNNs, thereby enhancing the performance of GNNs on a diverse range of graphs, ultimately matching or even surpassing these intricately designed GNN variants.

\subsection{LLMs for Graphs}
\label{sec:rw-llm}
The emergence of LLMs has inspired many explorations of utilizing LLMs for graph-structured data. Most of the works can be categorized into LLMs-as-Predictors and LLMs-as-Enhancers paradigms. 
\subsubsection{LLMs-as-Predictors}
Representative methods following this paradigm, e.g. NLGraph~\cite{wang2023can}, GPT4Graph~\cite{guo2023gpt4graph}, GraphQA~\cite{fatemi2023talk} and GraphText~\cite{zhao2023graphtext}, attempt to harness the zero-shot and few-shot ability, along with the in-context learning ability to solve graph tasks by describing the graph structural topology in natural language. However, the complex global graph structure can hardly be compressed in a token-limited prompt, thereby utilizing LLMs solely only achieves the performance far from desired. 
Besides directly prompting LLMs, GraphLLM~\cite{chai2023graphllm} conducts prefix tuning on an open-source LLM by concatenating graph-specific prefixes to the attention layers of a pre-trained LLM. InstructGLM~\cite{ye2023natural} and GraphGPT~\cite{tang2023graphgpt} adapt LLMs for graph downstream tasks through instruction tuning, employing natural language and a graph-text aligner to express graph structural information, respectively.
Nevertheless, neglecting the authenticated power of existing GNNs results in only moderate performance or substantial computing resource consumption.
\subsubsection{LLMs-as-Enhancers} 
Different from the former paradigm, LLMs-as-Enhancers paradigm incorporates LLMs to enhance input graphs before GNN training.
For example, TAPE~\cite{he2023harnessing} and Graph-LLM~\cite{chen2023exploring} prompt LLMs to interpret original texts attached to the nodes, and provide GNNs with semantically enhanced node features. SimTeG~\cite{duan2023simteg} parameter-efficiently finetunes (PEFT) LLMs with LoRA~\cite{hu2021lora} on the textual corpus of graphs and then use the finetuned LLM to generate node representations for GNN prediction.
LLM-TSE~\cite{sun2023large} explicitly instructs LLMs to produce a similarity score for two texts of two nodes, which subsequently leads to edge pruning. This methodology employs LLMs solely as a one-time data preprocessor before GNN training.
%In addition, inspired by how LLMs unify the natural language tasks, with a distinctive objective from ours, GraphPrompt~\cite{liu2023graphprompt}, Prodigy~\cite{huang2022towards}, All-in-one~\cite{sun2023large}, and OFA~\cite{liu2023one} aim to develop graph-structured prompt learning, which is analogous to natural language prompt learning, in order to further unify all downstream tasks in graph learning.

%\reminder{ruo yidian}
%In addition, there are also works~\cite{liu2023graphprompt,huang2022towards,sun2023large,liu2023one} in the field of graph prompting that apply the principles of LLMs to the domain of graph machine learning.
%Inspired by the route of how LLMs unify various natural language tasks, GraphPrompt~\cite{liu2023graphprompt}, Prodigy~\cite{huang2022towards}, All-in-one~\cite{sun2023large}, and OFA~\cite{liu2023one} aim to develop approaches for graph-structured prompt learning analogous to natural language prompt learning, with the objective of further unifying all downstream tasks in graph learning.

%\redfont{To summarize, the methods proposed within the LLMs-as-Predictors paradigm overlook the success of GNNs on graph-structured data, leading to suboptimal performance with higher resource consumption. The LLMs-as-Enhancers methods employ LLMs as data processors prior to GNN training, without adequately integrating the emerging intelligence of LLMs into the GNN training process.  
In summary, the LLMs-as-Predictors methods may result in suboptimal performance with higher resource consumption for overlooking the success of GNNs. The LLMs-as-Enhancers methods use LLMs to augment data before GNN training without integrating the LLM intelligence into GNN training. In comparison, our proposed LLM-as-Consultants paradigm aims to integrate LLMs into GNN training by consulting LLMs with a subset of nodes identified by GNNs as the most uncertain, in the attempt to enhance the classic GNNs with acceptable additional resource consumption. The core advancements of the LLMs-as-Consultants paradigm compared to previous paradigms, will be discussed detailedly in Section~\ref{sec:discussion}.

\section{LLMs-as-Consultants Paradigm}
\label{sec:discussion}
\begin{figure*}[htb]
  \centering
  \includegraphics[width=0.8\linewidth]{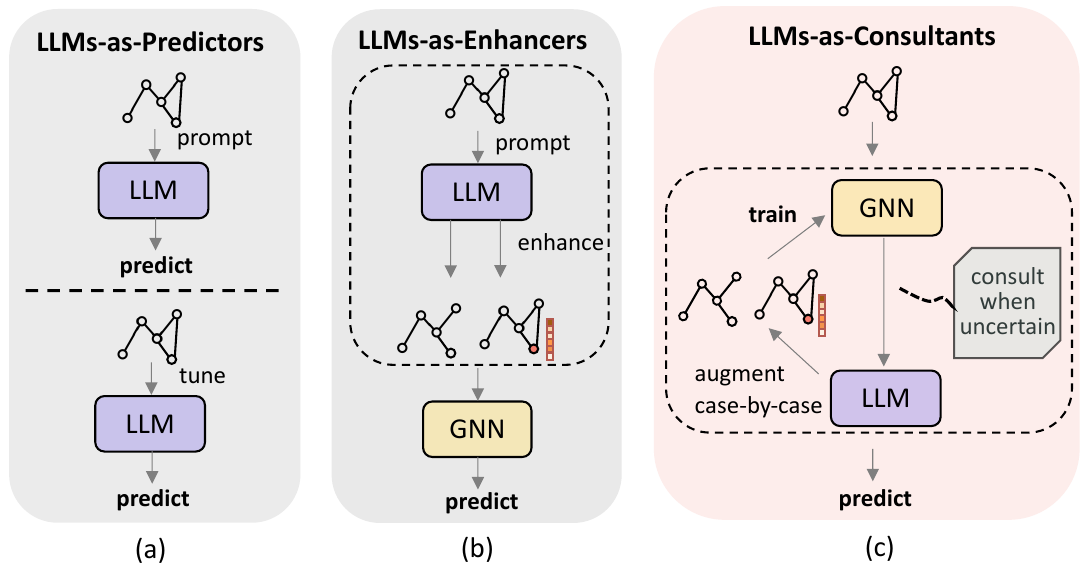}
  \caption{The paradigms for integrating LLMs with graphs.} 
  \label{fig:intro}
\end{figure*}

We display the pipelines of three paradigms for integrating LLMs with graphs in Fig.~\ref{fig:intro}.
 As shown in Fig.~\ref{fig:intro}~(a) and (b), some pioneering research introduced in Section~\ref{sec:rw-llm} has explored the potential of LLMs on graph data, which can be typically categorized into LLMs-as-Predictors and LLMs-as-Enhancers. 
 
 For the first kind, taking the node classification task as an example, some researchers harness the zero-shot and few-shot reasoning ability of LLMs to directly obtain node labels~\cite{zhao2023graphtext,fatemi2023talk,wang2023can,guo2023gpt4graph}. Others apply instruction tuning or prefix tuning to adapt LLMs specifically for graph tasks~\cite{chai2023graphllm,ye2023natural,tang2023graphgpt}. 
However, most of them disregard the advantages of GNNs, and tuning LLMs may require substantial computing resources as well.
For the latter one, LLMs-as-Enhancers, the LLMs are commonly utilized to enhance nodes' semantic features~\cite{chen2023exploring,he2023harnessing,duan2023simteg} or to refine local topological structures~\cite{sun2023large}. This kind of route is straightforward to apply LLMs in data preprocessing as a one-time enhancer, failing to couple the LLMs and GNNs interactively.

In contrast to existing approaches, our work introduces a novel paradigm that seamlessly integrates LLMs with GNNs through an interactive framework. We refer to this paradigm as "LLMs-as-Consultants" (see Fig.~\ref{fig:intro}~(c)). The fundamental distinction of our method lies in the recognition that LLMs can play an explicit and significant role in the training process of GNNs. This integration is not merely a superficial combination but an iteratively interactive manner where LLMs provide valuable insights and guidance that directly influence and enhance the learning capabilities of GNNs.

\section{Preliminaries}\label{sec:preliminary}
\subsection{Text-Attributed-Graphs~(TAGs)}
Text-attributed graphs ~(TAGs) are widely used in previous research on LLMs for graphs. A TAG can be formulated as:
\begin{equation}
    \mathcal{G}=\left(\mathcal{V}, \mathbf{A},\left\{s_n\right\}_{n \in \mathcal{V}}\right),
\end{equation}
where $\mathcal{V}$ is a set of $N$ nodes, $\mathbf{A} \in \{0,1\}^{N \times N}$ denotes the adjacency matrix, and $s_n \in \mathcal{D}^{L_n}$ is the text attached to node $n \in \mathcal{V}$, with $\mathcal{D}$ as the word dictionary, and $L_n$ as the text length. Note that our proposed framework is not limited to traditional TAGs that literally have texts as original attributes. In fact, most entities and their relations can be modeled and processed as graphs, and their characteristics can be expressed in text form.
\subsection{Graph Neural Networks~(GNNs)}
%% passing aggregation 
For GNN training, the texts associated with the nodes should be encoded to an embedded space. We represent the node embeddings as $\mathbf{X} \in \mathbb{R}^{N \times D}$, 
in which each row $\boldsymbol{}{x}_n$ denotes the corresponding node embedding, with $D$ as its dimension number.
%For node classification, GNNs aggregate information from a node’s neighbors, and then update the node representation with the aggregated information. 

A general GNN framework includes information aggregation and update processes~\cite{gilmer2017neural}. The aggregation function is a differentiable function for collecting and combining information from a node’s neighbors to produce a summary vector. It typically involves a form of pooling or summing the features from neighboring nodes. This process can be defined generically as follows:
%The k-th layer of a GNN can be formalized as:
\begin{equation}
\mathbf{x}_i^{(k)} = \text{AGGREGATE}^{(k)}\left( \left\{ \mathbf{x}_j^{(k-1)} : j \in \mathcal{N}(i) \right\} \right)
\end{equation}
Note that $x_i^{(k)} \in \mathbb{R}^{D}$ denotes the representation of node $i$ at layer $k$, with $\mathcal{N}(i)$ as the set of neighbors of node $i$.
After aggregating the features from the neighbors, the update function combines the aggregated features with the node’s own features from the previous layer. The formula for updating is as follows:
\begin{equation}
\mathbf{x}_i^{(k)} = \text{UPDATE}^{(k)}\left( \mathbf{x}_i^{(k-1)}, \mathbf{x}_i^{(k)} \right)
\end{equation}

%\begin{equation}
%x_i^{(l)} = f^{(l)}((\text{AGG}_{j \in \mathcal{N}(i)}^{(l)}x_j^{(l-1)}), x_i^{(l-1)}),
%\end{equation}
Aggregation and update are fundamental operations in GNNs, enabling nodes to exchange information and update their representations based on the graph structure and node features. These operations facilitate the learning of complex patterns and dependencies in graph-structured data. For node classification, the final node representations from the last layer are used as input to a classifier for predicting the class labels.

\subsection{Homophily Metrics}
In this subsection, we introduce homophily metrics defined as follows to distinguish homophilic and heterophilic graphs.

\begin{definition}[Graph-level Homophily Ratio]
\label{def:ghr} 
Given a graph $\mathcal{G} = (\mathcal{V}, \mathbf{A}, \mathbf{X})$ and the node labels $\mathbf{Y}$, graph-level homophily ratio $h_{\mathcal{G}}$ is defined as the fraction of intra-class edges: $h_{\mathcal{G}} = \frac{|\mathcal{E}_{\mathrm{intra}}|}{|\mathbf{A}|}$, where $\mathcal{E}_{\mathrm{intra}} = \{(u,v) \mid A_{uv} = 1 \wedge \mathbf{y}_u = \mathbf{y}_v\}$.
\end{definition}
\begin{definition}[Node-level Homophily Ratio]
\label{def:nhr} 
As to a node $v$ in $\mathcal{G}$, the node-level homophily ratio $h_{v}$ is defined as $h_{v} = \frac{| \mathcal{E}_v \cap \mathcal{E}_{\mathrm{intra}}|}{|\mathcal{E}_v|}$, where $\mathcal{E}_v = \{(u,v) \mid A_{uv} = 1\}$ is the set of edges linked to $v$. 
\end{definition}
%We adopt the above homophily metrics to categorize graph datasets and demonstrate our proposed L\scalebox{1}[0.8]{O}G\scalebox{1}[0.8]{IN} framework can achieve remarkable performance on graphs with varying degrees of homophily.
We utilize the aforementioned homophily metrics to categorize graph datasets and illustrate that our proposed L\scalebox{1}[0.8]{O}G\scalebox{1}[0.8]{IN} framework can help classic GNNs achieve remarkable performance on graphs exhibiting varying degrees of homophily.
\section{Methodology}\label{sec:method}
\begin{figure*}
  \centering
  \includegraphics[width=1.0\linewidth]{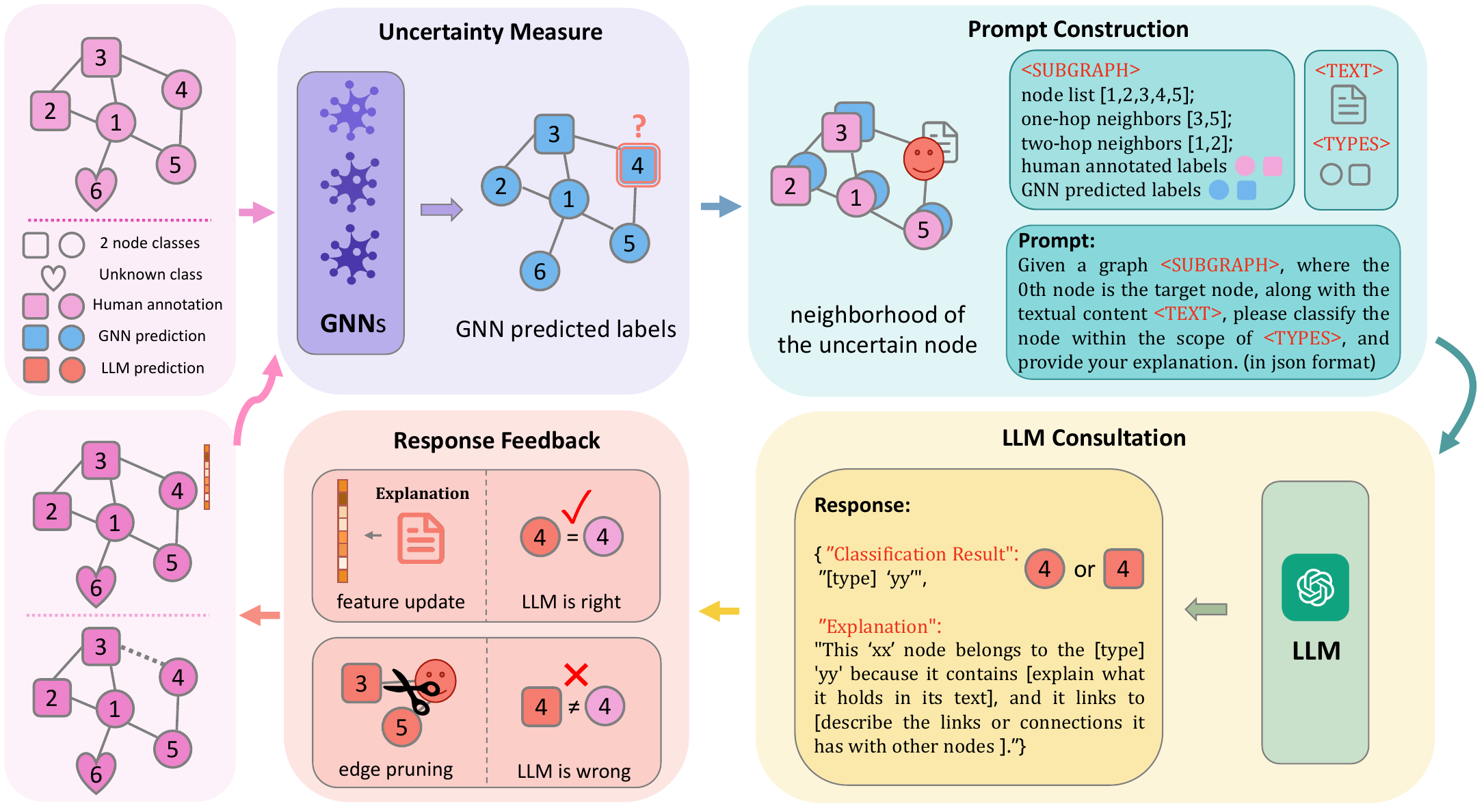}
  \vspace{-3mm}
  \caption{The pipeline of L\scalebox{1}[0.8]{O}G\scalebox{1}[0.8]{IN}.}
  \label{fig:LOGIN pipeline}
\end{figure*}
\subsection{Problem Formulation}
In this paper, we explore the integration of LLM consultation into GNN training for the node classification task. We follow the \textbf{transductive} setting of node classification: Given some labeled nodes $\mathcal{V_L} \subset \mathcal{V}$ in graph $\mathcal{G} = (\mathcal{V}, \mathbf{A} ,\left\{s_n\right\}_{n \in \mathcal{V}})$, we aim to classify the remaining unlabeled nodes $\mathcal{V_U}=\mathcal{V} \backslash \mathcal{V_L}$ within the same graph.
Formally, the target is to learn a set of GNN parameters $\mathbf{W}$ to predict the unlabeled nodes with the guidance from LLMs:
\begin{equation}
        %f_{\mathbf{W}|\text{LLM}}: (\mathbf{A, X}) \rightarrow \mathbf{Y}, ~ \left\{\mathbf{s}_n\right\}_{n \in \mathcal{V}}~\text{given}
        %$\mathcal{G} = (\mathcal{V}, \mathbf{A}, ,\left\{s_n\right\}_{n \in \mathcal{V}})$
        f_{\mathbf{W}|\text{LLM}}: (\mathbf{A, \left\{s_n\right\}_{n \in \mathcal{V}}}) \rightarrow \mathbf{Y}.
\end{equation}

Note that the ground truth labels of nodes, the pseudo labels predicted by GNNs, and the pseudo labels from LLMs are denoted as $\mathbf{Y},\mathbf{\hat{Y}}, \mathbf{\hat{Y}}^L \in \{0,1\}^{N \times C}$ respectively, with $C$ as the number of classes. 
%\redfont{It is noteworthy that LLMs only involve in training GNNs, and when testing we only use the GNNs trained with LLMs as consultants to predict.}

\subsection{Overview of L\scalebox{1}[0.8]{O}G\scalebox{1}[0.8]{IN}}%%TODO how to display
Figure~\ref{fig:LOGIN pipeline} demonstrates the framework of L\scalebox{1}[0.8]{O}G\scalebox{1}[0.8]{IN}. 
%In this pipeline, L\scalebox{1}[0.8]{O}G\scalebox{1}[0.8]{IN} seeks guidance from LLMs when GNNs are uncertain about some ambiguous nodes, and feeds responses from LLMs back to retrain GNNs with a complementary coping mechanism.
In this pipeline, we first identify target nodes that need consultation with LLM based on the GNN's predictive uncertainty~(c.f. Section~\ref{sec:method:node}). Then prompts consisting of the associated texts and local structures of these nodes are provided to LLMs for consultation~(c.f. Section~\ref{sec:method:llm}). After LLM consultation, a complementary mechanism is applied to fully utilize the LLMs' responses, regardless of their classification correctness~(c.f. Section~\ref{sec:method:response}). 

\subsection{Node Selection Based on Uncertainty} 
\label{sec:method:node}
An intuitive node selection approach is to consult the LLM with all the nodes on the graph. 
However, such a manner fails to utilize the capacity of GNNs, which are proficient at handling simple nodes, and it could be inefficient due to the limited interaction speed with LLMs. 
Hence, we decide to consult LLMs only with partial difficult nodes in the graph, and adopt the variance of the predictions from several GNNs as a surrogate indicator for identifying the hard ones in our L\scalebox{1}[0.8]{O}G\scalebox{1}[0.8]{IN}. 

Recall that Bayesian estimation is typically used to rate the predictive certainty in common neural networks~\cite{kendall2017uncertainties}.
Under the Bayesian framework, the conventionally fixed GNN parameters $\mathbf{W}$ are considered as random variables following specific distributions. The predictive probability of the Bayesian GNN with parameters $\mathbf{W_b}$ can be defined as  Eq.~\eqref{equ:pred prob}.
\begin{equation}
\label{equ:pred prob}
   p(\hat{\mathbf{Y}}\mid\mathbf{A}, \mathbf{X})=\int_{\mathbf{W_b}} p(\hat{\mathbf{Y}} \mid \mathbf{W_b}, \mathbf{A}, \mathbf{X}) p(\mathbf{W_b} \mid \mathbf{A}, \mathbf{X}) \mathrm{d} \mathbf{W_b}
\end{equation}

Since the true posterior $p(\mathbf{W} \mid \mathbf{A}, \mathbf{X})$ in Eq.~\eqref{equ:pred prob} is hard to calculate in practice, variational inference uses an arbitrary distribution $q_{\theta}(\mathbf{W})$ to approximate the posterior. Specifically, we use MC dropout variational inference, in which dropout could serve to perform variational inference. In this method, the variational distribution $q_{\theta}(\mathbf{W})$ is from a Bernoulli variable $\mathbf{M}_{\theta}$, representing whether the neurons are on or off, as Eq. \eqref{equ:var dist} shows.
%%In MC dropout variational inference, the variational distribution $q_{\theta}(\mathbf{W})$ is from a Bernoulli variable $\mathbf{M}_{\theta}$, with $\theta$ as the dropout rate in neural networks, which can be defined as Eq. \eqref{equ:var dist},

\begin{equation}
\begin{aligned}
\label{equ:var dist}
    %%\textbf{M}_{\theta} \sim \operatorname{Bernoulli}(\theta),\qquad 
\mathbf{M}_{\theta} & {\sim \rm{Bernoulli}(\mathbf{\theta})},\\
q _ {\theta}(\mathbf{W})&=p(\mathbf{W}\mid\theta)=p(\mathbf{W} _ \theta),\\
\mathbf{W}_{\theta} &= \mathbf{M}_{\theta} \odot \mathbf{W} _ b.
\end{aligned}
\end{equation}

where $\theta$ is the dropout rate, and $\mathbf{M} _ {\theta}$ represents a binary mask that controls which neurons in GNNs are off.
%with $\theta$ and $\mathbf{W}_{mc}$ here as variational parameters. Noted $\mathbf{M}_{\theta}$ also represents a binary mask that controls which neurons in GNNs are deactivated during training with the dropout technique.
Then, $\{ \hat{\mathbf{W}}_t \}_{t=1}^T$ is a collection of $T$ samples drawn from $\mathbf{W}_{\theta}$ in a Monte Carlo way. Through the minimization of the loss function defined in Eq.~\eqref{equ:ploss} using these weight samples, $\mathbf{W}_{b}$ can be acquired. 
\begin{equation}
\label{equ:ploss}
\mathcal{L}(\mathbf{W}_{b})=-\frac{1}{T} \sum_{t=1}^{T} \mathbf{Y} \log \left( f_{\hat{\mathbf{W}}_t}(\mathbf{A}, \mathbf{X}) \right) + \frac{1-\theta}{2 T}\|\mathbf{W}_{b}\|^{2}
\end{equation}

After $\mathbf{W} _ {b}$ is trained,  the model uncertainty score $\mathbf{U}$, which is an $N$-dimension vector indicating the uncertainty score of each node, is calculated as in Eq.~\eqref{equ:model:uncertainty}. 
%In the end, the predictive uncertainty of the trained model is defined as the variance of its predictions. This uncertainty vector can be straightforwardly computed using the formula in Eq.~\eqref{equ:model:uncertainty}, in which the dimension with the highest score corresponds to the most uncertain node of GNN.
\begin{equation}
\label{equ:model:uncertainty}
\begin{aligned}
U(\hat{\mathbf{Y}} \mid \mathbf{A}, \mathbf{X})  &=\operatorname{Var}(\hat{\mathbf{Y}} \mid \mathbf{A}, \mathbf{X}) \approx \frac{1}{T} \sum_{t=1}^{T} \left( \hat{Y}_{t} - \frac{1}{T} \sum_{t=1}^{T} \hat{Y}_{t}  \right)^2
\end{aligned}
\end{equation}

%% TODO revise captions in fig1
\subsection{LLM Consultation} 
\label{sec:method:llm}
After we identify the nodes with the most uncertainty $\mathcal{V}_{uc}$ for LLM consultation, the next step in L\scalebox{1}[0.8]{O}G\scalebox{1}[0.8]{IN} is to figure out how to convert the semantic and topological information of these nodes into an understandable format by LLMs. 
We consult LLMs in a zero-shot node-by-node manner, i.e., constructing a single prompt for each uncertain node, without classification examples provided. 
Through this way, no fine-tuning method is involved, and the parameters of LLMs are fixed. Each node prompt is composed of three elements: instruction, input data, and output indicator. The Prompt Construction module in Fig.~\ref{fig:LOGIN pipeline} shows a general prompt example, and we provide the detailed prompts in Appendix \ref{sec:app:prompts}.

\subsubsection{Instruction.}The instruction explains the node classification task in the context of the graph type, depicting real-world meanings of nodes and edges, with a spectrum of the categories provided.

\subsubsection{Input Data.} The input data includes information related to the target node $n$, namely its original text $s_n$, the two-hop neighborhood $\mathcal{N}_2(n)$ description, and the neighbor labels $\mathbf{Y_{\mathcal{N}_2(n)}},\mathbf{\hat{Y}_{\mathcal{N}_2(n)}}$ from both human annotations and GNN predictions. Instead of listing edges to express connectivity~\cite{chen2023exploring,guo2023gpt4graph}
, we opt to summarize one-hop and two-hop neighbors for the sake of semantic clarity and easier parsing. Note that we only select uncertain nodes from the train set, otherwise, it might bring a data leakage. The target node's labels $y_n,\hat{y}_n$ are also  included in its corresponding prompt to provide more information for LLM predictions.

\subsubsection{Output Indicator.}The output indicator is used to control the output format of LLMs for further analysis, specifically by offering a desired response example in JSON format.  Regarding the response content, we aim for the output to include the most probable classification outcome $\hat{y}_n^L$ along with its explanation $e_n$. 

\subsection{LLM Response Feedback}
\label{sec:method:response}
After consulting LLMs with all selected uncertain nodes $\mathcal{V}_{uc}$, we have a group of LLM predicted pseudo labels and corresponding explanations $ \left\{{n \in \mathcal{V}_{uc}} : (\hat{y}_n^L, e_n)\right\}$. Next, we aim to maximize the utilization of LLM responses and convert the information into signals suitable for GNN processing.
%We distinguish between right and wrong LLM responses using ground truth labels and offer distinct approaches to fully leverage the positive and negative data.
The LLM responses can be divided into two types compared to ground truth labels, i.e., the correct predictions when $\hat{y}_n^L = y_n$ and the wrong predictions when $\hat{y}_n^L \neq y_n$. We not only utilize the correct predictions but also the misclassifications, and offer distinct approaches respectively. Incorporating such a complementary coping mechanism endows GNN training with full exploitation of LLM responses. 

For clarity in expression, we denote the correct nodes and incorrect nodes as $\mathcal{V}_{r}$ and $\mathcal{V}_{w}$ respectively, satisfying $\mathcal{V}_{r} \cup \mathcal{V}_{w} = \mathcal{V}_{uc}$ and $\mathcal{V}_{r} \cap \mathcal{V}_{w} = \emptyset$. 

%Wrong answers from LLMs are confirmed to contain useful knowledge as well~\cite{li2023turning}. Inspired by this insight, 

\subsubsection{When LLM is Right.}
%For a right node $n \in \mathcal{V}_r$, we update its original node embedding $\mathbf{x}_n$ with the embedding of the corresponding explanation $e_n$ parsed from the LLM response:
For a right node $n \in \mathcal{V}_r$, we append the explanation $e_n$ parsed from the LLM response to its original text $s_n$. And we attain its new node embedding $\mathbf{{x}_n}'$ by encoding the new attached text from $e_n$:
\begin{equation}
\label{equ:update fea}
    \left\{\mathbf{{x}_n}'\right\}_{n \in \mathcal{V}_r} \leftarrow \text{ENC}\left(\left\{s_n + e_n\right\}_{n \in \mathcal{V}_r})\right.
\end{equation}

\subsubsection{When LLM is Wrong.}
\label{sec:sycophancy}
For those wrong nodes, the LLM responses are left out since the explanations for incorrect classifications could be unhelpful.
%Given the emerging reasoning ability that LLMs have shown in natural language tasks~\cite{brown2020language},
Instead, we opt to attribute the misclassification to the topological information, e.g., complex structure patterns and potential adversarial noises. Therefore, we heuristically decide to simplify the local structure for the wrong nodes to make it easier for GNNs to learn.
%We leverage the sycophancy of LLMs, that the models' responses are adversely affected by spurious correlations in the input~\cite{weston2023system}, to determine which node's ego-graph contains misleading links that result in the wrong answer. 
%We determine which node's ego-graph contains misleading links that result in the wrong answer
Specifically, around a wrong node $n \in \mathcal{V}_w$, we prune edges based on node similarity scores to denoise the local structure. 
As in representative graph structure learning methods, e.g. GNNGuard~\cite{zhang2020gnnguard}, we measure similarity $d_{ni}$ between the features of the wrong node $n$ and its neighbor $i$ using cosine similarity:
\begin{equation}
\label{equ:sim}
    d_{ni} = d^{cos}(\boldsymbol{x}_n,~\boldsymbol{x}_i) = \left(\boldsymbol{x}_n \odot \boldsymbol{x}_i\right) /\left(\left\|\boldsymbol{x}_n\right\|_2\left\|\boldsymbol{x}_i\right\|_2\right).
\end{equation}
Then, edges of node $n$ are pruned if their similarity scores fall below a user-defined threshold $d_{th}$. By pruning edges of the misclassified nodes $\mathcal{V}_w$ as in Eq.~\ref{equ:pruning}, we accomplished the topological refinement leveraging the wrong answers from LLMs.
\begin{equation}
\label{equ:pruning}
    \mathbf{A'} \leftarrow \mathbf{1}{\left(\mathbf{D}  - d_{th}\cdot\mathbf{I}\right)} \odot \mathbf{M}_w \odot \mathbf{A} \\
\end{equation}
Here $\mathbf{D}=\left(d_{ij}\right)_{\mathbf{N} \times \mathbf{N}}$ denotes the similarity matrix, with $\mathbf{M}_w$ as a binary node mask representing $\mathcal{V}_w$, and $\mathbf{1}$ is a binary indicator.
%\begin{equation}
%\label{equ:indicator}
    %\mathbf{1} (d_{ij}-d_{th}) = \begin{cases} 1 & \text{if}~{(d_{ij}-d_{th})>0}, \\ 0 & \text{otherwise}. \end{cases}
%\end{equation}

In a nutshell, the response feedback stage is divided into two complementary scenarios. When the LLM classifies nodes correctly, we upgrade the original node embeddings with the encoded explanations. Conversely, we impute the misclassifications to the topological noise, hence pruning edges around the wrong nodes. 

\subsection{Train and Test}
%In L\scalebox{1}[0.8]{O}G\scalebox{1}[0.8]{IN} pipeline, we retrain GNNs after the LLM response feedback stage. 
After the LLM response feedback stage, L\scalebox{1}[0.8]{O}G\scalebox{1}[0.8]{IN} is set to retrain GNN. When retraining, the GNN aggregates the original node features and the ones updated with the correct classification grounds. And for the nodes that LLMs misclassify, aggregation only happens along the retained edges after structure refinement. In this way, the message passing of GNNs is integrated with the LLM intelligence both semantically and topologically. %This integration enables L\scalebox{1}[0.8]{O}G\scalebox{1}[0.8]{IN} to reach or even surpass the methods which also enhance the testing nodes with LLMs. 

The GNN training process with one-time consultation within the L\scalebox{1}[0.8]{O}G\scalebox{1}[0.8]{IN} framework is illustrated in Algorithm~\ref{alg:login}. This process can be iteratively repeated until the maximum number of iterations is reached or the performance of the GNN has achieved an acceptable level. 
%可以问多轮，gnn可以按照如上流程重训练，达到迭代次数上限，GNN不断加强，测试时没有label，不需要咨询大模型，直接用gnn做分类
 %We will further prove the outperformance of L\scalebox{1}[0.8]{O}G\scalebox{1}[0.8]{IN} empirically in the Section~\ref{sec:exp}.
 
During the test, the ground truth labels are inaccessible, so we only use the trained GNN without additional LLM consultation to make final predictions. Since we retrain the GNN over the graphs augmented by LLM consultation, i.e., with the node representations enhanced and the noisy links removed, the attained GNNs can perform aggregation in a more reliable neighborhood with more accurate semantic information and thus produce improvements.
%While LLMs present impressive performance advantages, their widespread adoption is hindered by significant computational resource requirements. Recall that our motivation for proposing the LLMs-as-Consultants paradigm is to achieve a balance, enhancing classic GNNs with tolerant resource consumption. In addition to this, to ensure fairness and prevent data leakage, it's worth noting that we conduct node prediction inference in a non-LLM manner, i.e., we do not consult LLMs when testing. 
%It is noteworthy that LLMs only involve in training GNNs, and when testing we only use the GNNs trained with LLMs as consultants to predict.
%% retrain GNN

\begin{algorithm}[h]
%\iffalse 
\caption{L\scalebox{1}[0.8]{O}G\scalebox{1}[0.8]{IN}}
\label{alg:login}
\KwIn{$\mathcal{G} = (\mathcal{V}, \mathbf{A}, ,\left\{s_n\right\}_{n \in \mathcal{V}})$: A TAG, $\mathcal{V}_{\mathrm{train}}$: Set of training nodes, $\mathbf{X}$: Node representations, $\mathbf{Y}$: Labels of training nodes, $T$: Number of Monte Carlo samples, $\theta$: Dropout rate for uncertainty estimation, $\gamma$: Ratio of uncertain nodes, $d_{th}$: Similarity threshold}
\KwOut{The retrained GNN parameters after LLM consultation $\mathbf{W}_r$. }

    Choose basic models for pretrained GNN $f_{\mathbf{W}_p}$ and retrained GNN $f_{\mathbf{W}_r}$; \label{alg:choose}\\
    
    Initialize parameters $\mathbf{W}_p$ and $\mathbf{W}_r$; \label{alg:init} \\
    
    \For {$t = 1, \ldots, T$} { \label{alg:loop-begin}
        Sample $\hat{\mathbf{W}}_t$ from $\mathbf{W_\theta}$ 
    }
    Train $\mathbf{W}_p$ and $\{\hat{\mathbf{W}}_t\}_{t=1}^T$ by minimizing Eq.~(\ref{equ:ploss}); \\ \label{alg:train:b}
    Estimate the model uncertainty $U$ w.r.t. Eq.~(\ref{equ:model:uncertainty}); \\ \label{alg:score}
    Divide $\mathcal{V}_{\mathrm{train}}$ into $\mathcal{V}_{\mathrm{c}}$ and $\mathcal{V}_{\mathrm{uc}}$ according to $U$ such that $\gamma = \frac{|\mathcal{V}_{\mathrm{uc}}|}{|\mathcal{V}_{\mathrm{train}}|}$; \\ \label{alg:divide}
    \For{$n \in \mathcal{V}_{\mathrm{uc}}$}{
    Consult the LLM with prompt $n$ and gain LLM-predicted label and its explanation~$\hat{y_n}^L, e_n$ ;\\

        \eIf{$\hat{y_n}^L = y_n$}{
    Update node features $\boldsymbol{x}_n$ w.r.t. Eq.~(\ref{equ:update fea});\\}
    {Prune edges of $n$ w.r.t. Eq.~(\ref{equ:pruning});\\}
      }
     Retrain GNN $f_{\mathbf{W}_r}$ with new $\mathbf{A'}, \mathbf{X'}$;\\
     Return $\mathbf{W}_r$.
\end{algorithm}
\section{Evaluation}\label{sec:exp}
In this section, we investigate the effectiveness of our L\scalebox{1}[0.8]{O}G\scalebox{1}[0.8]{IN} framework on both homophilic and heterophilic graph datasets, to address the following research questions:

\begin{itemize}
\item \textbf{RQ1}: Does the L\scalebox{1}[0.8]{O}G\scalebox{1}[0.8]{IN} framework achieve performance comparable to state-of-the-art GNNs?
\item \textbf{RQ2}: How does the complementary coping mechanism for LLMs' responses contribute to the L\scalebox{1}[0.8]{O}G\scalebox{1}[0.8]{IN} framework?
\item \textbf{RQ3}: How does L\scalebox{1}[0.8]{O}G\scalebox{1}[0.8]{IN} operate over specific nodes based on responses from LLMs?
\item \textbf{RQ4}: Can consulting more advanced LLMs in L\scalebox{1}[0.8]{O}G\scalebox{1}[0.8]{IN} unlock greater potential?
\item \textbf{RQ5}: How do models under the LLMs-as-Consultants paradigm perform compared with LLMs-as-Predictors and LLMs-as-Enhancers?
\item \textbf{RQ6}: Can consulting LLMs with more training nodes achieve higher performance increases with classic GNNs?
\item \textbf{RQ7}: Can L\scalebox{1}[0.8]{O}G\scalebox{1}[0.8]{IN} also help advanced GNNs enhance performance?
\end{itemize}

\subsection{Experimental Setup}
\subsubsection{Datasets}
We conducted extensive experiments on six datasets: three homophilic graphs and three heterophilic graphs, to demonstrate the versatility and applicability of our proposed L\scalebox{1}[0.8]{O}G\scalebox{1}[0.8]{IN} framework in handling graphs with distinct characteristics. For homophilic graphs, we collected the Cora~\cite{mccallum2000automating} and PubMed~\cite{sen2008collective} datasets from widely used TAG benchmarks, while Arxiv-23~\cite{he2023harnessing} was recently introduced to eliminate the data leakage unfairness when evaluating the impact of LLMs on graph learning. 

For heterophilic graphs, we transformed the commonly-used web-page-link graphs: Wisconsin, Texas, and Cornell~\cite{craven1998learning} into TAGs by sourcing and incorporating the raw texts\footnote{\url{http://www.cs.cmu.edu/~webkb/}}, which were not available previously in graph libraries. Besides, we fine-tuned DeBERTa-base~\cite{he2021deberta} to encode raw texts into node embeddings. The basic statistics of the datasets are displayed in Table~\ref{tab:exp:data}. The proportions of heterophilic nodes in graphs are calculated via Definition~\ref{def:nhr}: If the node-level homophily score of node $n$ is below 0.5, then $n$ is defined as a heterophilic node.
\begin{table}[htbp]
  \centering
  \caption{Statistics of the datasets.}
  \vspace{-2mm}
    \begin{tabular}{c|rrccc}
    \toprule
    Dataset & \multicolumn{1}{c}{\#Node} & \multicolumn{1}{c}{\#Edge} & \#Class & \#Feat & \%Heter \\
    \midrule
    Cora  & 2708  & 5429  & 7    & 768  & 12\% \\
    PubMed & 19717 & 44338 & 3 & 768 & 16\% \\
    Arxiv-23 & 46198  & 78548  & 40    & 768  & 8\% \\
    Wisconsin & 265  & 938  & 5    & 768   & 79\% \\
    Texas & 187  & 578  & 5    & 768   & 91\% \\
    Cornell & 195  & 569  & 5    & 768   & 88\% \\
    \bottomrule
    \end{tabular}%
  \label{tab:exp:data}%
\end{table}%
\subsubsection{Compared Methods}
We integrate fundamental GNNs with L\scalebox{1}[0.8]{O}G\scalebox{1}[0.8]{IN}, namely GCN~\cite{kipf2016semi}, GraphSAGE~\cite{hamilton2017inductive} and MixHop~\cite{abu2019mixhop}, to compare with the advanced state-of-the-art GNNs, including JK-Net~\cite{xu2018representation}, H2GCN~\cite{zhu2020beyond}, APPNP~\cite{gasteiger2018predict}, GCNII~\cite{chen2020simple}, SGC~\cite{wu2019simplifying} ,and SSP~\cite{izadi2020optimization}, to demonstrate that the former can achieve performance on par with the latter. MLP, which predicts without adjacency information, also serves as a base learner combined with L\scalebox{1}[0.8]{O}G\scalebox{1}[0.8]{IN} to show the potential of our framework.
We select Vicuna-v1.5-7b as our LLM consultant, an open-source LLM trained by fine-tuning Llama 2 on user-shared conversations.
%Additionally, we showcase that LLMs alone achieve only moderate performance, highlighting the necessity of our case-by-case processing mechanism to analyze LLMs' responses. 

For MLP, GCN, GraphSAGE, and MixHop, ``OR'' stands for the vanilla backbone model trained with original shallow node features, and ``FT'' denotes the vanilla backbone utilizing node embeddings encoded by the fine-tuned LM. ``LO'' represents the model trained within our L\scalebox{1}[0.8]{O}G\scalebox{1}[0.8]{IN} framework. Besides, in the ablation study, we refer to the feature update and structure refinement operations in the LLM response feedback stage as ``F'' and ``S'', respectively.

In addition to advanced GNNs, we also compared L\scalebox{1}[0.8]{O}G\scalebox{1}[0.8]{IN} as an implementation of LLMs-as-Consultants paradigm, with the other existing paradigms, i.e. LLMs-as-Predictors and LLMs-as-Enhancers, respectively.

\subsubsection{Implementations}
We adopt \textbf{accuracy} on the test set of the nodes to evaluate the node prediction performance of all the listed models. We report the mean accuracy and standard error from five runs with varied random data splits, and all the reported results are statistically significant.

For hyper-parameters, in the uncertainty measure module, we implement Monte Carlo dropout variational inference by running models $T$ times with different neurons off, where $T$ represents the number of weight samples in MC dropout. $T$ is always set to 5 in our experiments. For the ratio $\gamma$ of uncertain nodes to consult, we adjust $\gamma$ slightly on each dataset around the proportion of heterophilic nodes shown in Table~\ref{tab:exp:data}. In the response feedback stage, we tune the similarity threshold $s_{th}$ between $[0.1,0.2]$, according to the off-shell tool GNNGuard~\cite{zhang2020gnnguard}.

\subsection{Performance Comparison~(RQ1)}

\begin{table*}[htbp] %htbp
\small
  \centering
  \caption{Performance comparison for node classification on homophilic and heterophilic graphs. The best results and the second best results among the classic GNNs with and without L\scalebox{1}[0.8]{O}G\scalebox{1}[0.8]{IN} and the advanced models are bold and underlined respectively.}
    \begin{tabular}{c|cc|c|c|c|c|c|c}
    \toprule
    \multicolumn{3}{c|}{Method} & \multicolumn{1}{c|}{Cora} & \multicolumn{1}{c|}{PubMed} & \multicolumn{1}{c|}{Arxiv-23} &
    \multicolumn{1}{c|}{Wisconsin} & \multicolumn{1}{c|}{Texas} &
    \multicolumn{1}{c}{Cornell} \\
    \midrule
    %%\shortstack{Non-\\Specialized}
 \multirow{12}{*}{\shortstack{Classic\\GNNs}}
   & \multirow{3}[3]{*}{MLP}
     & \multicolumn{1}{l|}{OR} &  0.6438 $\pm$ 0.0331 & 0.8805 $\pm$ 0.0032  & 0.6759 $\pm$ 0.0027  & 0.8113 $\pm$ 0.0718 & 0.8105 $\pm$ 0.0730 & 0.7538 $\pm$ 0.0669 \\
   & & \multicolumn{1}{l|}{FT} & 0.6897 $\pm$ 0.0102 & 0.9486 $\pm$ 0.0030  & 0.7789 $\pm$ 0.0023  & \underline{0.8415 $\pm$ 0.0391} & \underline{0.8211 $\pm$ 0.0681}  & \underline{0.8049 $\pm$ 0.0602}  \\
         & & \multicolumn{1}{l|}{LO} &  0.7063 $\pm$ 0.0201 & 0.9505 $\pm$ 0.0036  & \underline{0.7902 $\pm$ 0.0034}  & \textbf{0.8528 $\pm$ 0.0588} & \textbf{0.8895 $\pm$ 0.0820} & \textbf{0.8051 $\pm$ 0.0618} \\
        \cmidrule{2-9}  

   &\multirow{3}[3]{*}{GCN}
       &\multicolumn{1}{l|}{OR} &  0.8630 $\pm$ 0.0219 & 0.8635 $\pm$ 0.0083  & 0.6707 $\pm$ 0.0040 & 0.3736 $\pm$ 0.0672 & 0.4579 $\pm$ 0.0711 & 0.4308 $\pm$ 0.0664\\
    && \multicolumn{1}{l|}{FT} & 0.8683 $\pm$ 0.0191 & 0.9289 $\pm$ 0.0069  & 0.7624 $\pm$ 0.0051  & 0.4415 $\pm$ 0.1152 & 0.5526 $\pm$ 0.0832  & 0.5282 $\pm$ 0.0644  \\
        &&  \multicolumn{1}{l|}{LO} & 0.8694 $\pm$ 0.0177 & 0.9396 $\pm$ 0.0030  & 0.7703 $\pm$ 0.0020 &0.5057 $\pm$ 0.0430  & 0.5789 $\pm$ 0.0588 & 0.5231 $\pm$ 0.0292  \\
              \cmidrule{2-9}

   & \multirow{3}[3]{*}{GraphSAGE} 
       & \multicolumn{1}{l|}{OR} & \underline{0.8720 $\pm$ 0.0216} & 0.8849 $\pm$ 0.0026  & 0.6864 $\pm$ 0.0011  & 0.6113 $\pm$ 0.0662 & 0.5053 $\pm$ 0.0776 & 0.6051 $\pm$ 0.0389 \\
    && \multicolumn{1}{l|}{FT} &0.8592 $\pm$ 0.0363 & 0.9472 $\pm$ 0.0026 & 0.7881 $\pm$ 0.0019  & 0.7211 $\pm$ 0.1324 & 0.7579 $\pm$ 0.1123 &  0.7179 $\pm$ 0.1189 \\
          && \multicolumn{1}{l|}{LO} &\textbf{ 0.8727 $\pm$ 0.0219 }& \underline{0.9511 $\pm$ 0.0036}  & \textbf{0.7941 $\pm$ 0.0029}  & 0.7434 $\pm$ 0.0930 & 0.7737 $\pm$ 0.1311 &  0.6872 $\pm$ 0.0896 \\
     \cmidrule{2-9}  
    &\multirow{3}[3]{*}{MixHop} 
       & \multicolumn{1}{l|}{OR} &   0.8601 $\pm$ 0.0281 & 0.8969 $\pm$ 0.0038  & 0.6774 $\pm$ 0.0029  & 0.5736 $\pm$ 0.1183 & 0.5526 $\pm$ 0.1500 & 0.4974 $\pm$ 0.0803 \\
  &  & \multicolumn{1}{l|}{FT} & 0.8572 $\pm$ 0.0123 & 0.9493 $\pm$ 0.0030  & 0.7775 $\pm$ 0.0036  & 0.7092 $\pm$ 0.1035 & 0.7421 $\pm$ 0.1075  & 0.6718 $\pm$ 0.1397  \\
        &  & \multicolumn{1}{l|}{LO} & 0.8624 $\pm$ 0.0253 & \textbf{0.9513 $\pm$ 0.0038} & 0.7818 $\pm$ 0.0040  & 0.7094 $\pm$ 0.0738 & 0.8158 $\pm$ 0.0930 & 0.7179 $\pm$ 0.0314 \\

    \midrule
       \multirow{6}{*}{\shortstack{Advanced\\GNNs}}
    &\multicolumn{2}{c|}{JK-Net} & 0.8579 $\pm$ 0.0001 & 0.8841 $\pm$ 0.0001  & 0.7532 $\pm$ 0.0012 & \underline{0.7431 $\pm$ 0.0041} & 0.6649 $\pm$ 0.0046 & 0.6459 $\pm$ 0.0075 \\
    &\multicolumn{2}{c|}{H2GCN} & \underline{0.8692 $\pm$ 0.0002} & 0.8940 $\pm$ 0.0001  & 0.7382 $\pm$ 0.0011  & \textbf{0.8667 $\pm$ 0.0022}& \textbf{0.8486 $\pm$ 0.0044} & \textbf{0.8216 $\pm$ 0.0023}  \\
    &\multicolumn{2}{c|}{APPNP} & 0.8539 $\pm$ 0.0477 & \textbf{0.9355 $\pm$ 0.0060} & \underline{0.7969 $\pm$ 0.0143} & 0.6830 $\pm$ 0.0470 &\underline{0.7368 $\pm$ 0.0832}  & 0.6410 $\pm$ 0.0480  \\
    &\multicolumn{2}{c|}{GCNII} & \textbf{0.8833 $\pm$ 0.0027} & 0.7925 $\pm$ 0.0043  & 0.7847 $\pm$ 0.0068  & 0.7020 $\pm$ 0.0037 & 0.7135 $\pm$ 0.0039 & \underline{0.7405 $\pm$ 0.0060}  \\ 
    &\multicolumn{2}{c|}{SGC} & 0.8509 $\pm$ 0.0648 & 0.8832 $\pm$ 0.0055  & 0.7740 $\pm$ 0.0160  & 0.5321 $\pm$ 0.0506 & 0.5526 $\pm$ 0.0811 & 0.4615 $\pm$ 0.0748  \\
   & \multicolumn{2}{c|}{SSP} & 0.8616 $\pm$ 0.0289 & \underline{0.9178 $\pm$ 0.0116}  & \textbf{0.7976$\pm$ 0.0185}  & 0.6302 $\pm$ 0.0850 & 0.7000 $\pm$ 0.0758 & 0.6923 $\pm$ 0.0314  \\
    %\midrule
   %% \multicolumn{2}{c|}{Vicuna-v1.5-7b} & 0.8926 $\pm$ 0.0085 & 0.9178 $\pm$ 0.0116  & 0.7955 $\pm$ 0.0185  & 0.6302 $\pm$ 0.0850 & 0.7000 $\pm$ 0.0758 & 0.6923 $\pm$ 0.0314  \\
   %% \multicolumn{2}{c|}{Vicuna-v1.5-13b} & 0.8926 $\pm$ 0.0085 & 0.9178 $\pm$ 0.0116  & 0.7955 $\pm$ 0.0185  & 0.6302 $\pm$ 0.0850 & 0.7000 $\pm$ 0.0758 & 0.6923 $\pm$ 0.0314  \\
    \bottomrule
    \end{tabular}%
  \label{tab:exp:main}%
\end{table*}%

To answer the first research question, we evaluate our proposed L\scalebox{1}[0.8]{O}G\scalebox{1}[0.8]{IN} framework on three homophilic datasets and three heterophilic datasets. The prediction accuracy scores and standard errors are reported in Table~\ref{tab:exp:main}. We have the following observations.

\subsubsection{Comparison with Vanilla Baselines.} Firstly, our method consistently outperforms the vanilla baselines trained with the original node features or the LM-finetuned node embeddings in most cases. There are only two exceptions in the Cornell dataset. Since Cornell is a relatively small dataset with only 39 nodes in the test set, as indicated by the standard errors, the experimental randomness is quite high with this small dataset. Nevertheless, L\scalebox{1}[0.8]{O}G\scalebox{1}[0.8]{IN} still helps improve the performance of MixHop on Cornell by 4.6\% with a low deviation. Apart from this exception, all listed fundamental baselines trained within our L\scalebox{1}[0.8]{O}G\scalebox{1}[0.8]{IN} framework exceed the vanilla ones in node classification accuracy on the other five datasets, demonstrating the effectiveness of integrating LLMs as consultants into the GNN training process.

\subsubsection{Comparison with Advanced GNNs.} Secondly, we are able to attain performance comparable to that of advanced GNNs by training fundamental models within the L\scalebox{1}[0.8]{O}G\scalebox{1}[0.8]{IN} framework. It is noteworthy that on PubMed and Texas, respectively known as benchmarks for homophilic and heterophilic graphs, we achieve the highest prediction accuracy among all the compared methods. This finding verifies the generalizability of our method on graphs with distinctive characteristics. For Cora, Arxiv-23, Wisconsin, and Cornell, L\scalebox{1}[0.8]{O}G\scalebox{1}[0.8]{IN} achieve remarkable performance on par with the SOTA GNNs such as H2GCN and SSP as well.

\subsubsection{Comparison with the Simplest Model.} Thirdly, it draws our attention that regardless of the feature type and training paradigm we apply, MLP reveals great potential in the node classification task on heterophilic graphs. We believe that the TAGs with considerably high heterophily may be better regarded as natural language data rather than graph-structured data, for the links contained in these datasets seem to only carry 
noticeably limited information. Additionally, our L\scalebox{1}[0.8]{O}G\scalebox{1}[0.8]{IN} can clearly enhance MLP, inspiring further research on how to leverage simple existing tools to achieve superb capability with the help of the LLM-as-Consultants paradigm.

\subsection{Ablation Study~(RQ2)}

To answer the second research question, we subdivide the response feedback stage into two distinct components, namely feature update and structure refinement. To show both of them contribute to our complementary coping mechanism for utilizing responses from LLMs, we remove these two modules respectively.
%In this subsection, we aim to show our complementary coping mechanism combining these two components is essential for the displayed performance improvements.

We present the results of ablation studies exclusively on the PubMed and Texas datasets, where L\scalebox{1}[0.8]{O}G\scalebox{1}[0.8]{IN} has demonstrated superior performance, as depicted in Fig.~\ref{fig:ablation}. Across both datasets, the complete L\scalebox{1}[0.8]{O}G\scalebox{1}[0.8]{IN} pipeline consistently achieves the highest performance, thus affirming the effectiveness and necessity of our complementary design for processing LLMs' responses.

\subsubsection{Feature Update.} Regarding the feature update component, we observe that it enhances prediction accuracy more effectively for a homophilic graph. Specifically, in the case of Cora, we notice from Fig.~\ref{fig:ablation}~(a) that L\scalebox{1}[0.8]{O}G\scalebox{1}[0.8]{IN} without feature updates results in a notable decrease in performance on certain occasions compared to the whole pipeline.

\subsubsection{Structure Refinement.} 
\begin{figure}[!htbp]
  \centering
  \includegraphics[width=\linewidth]{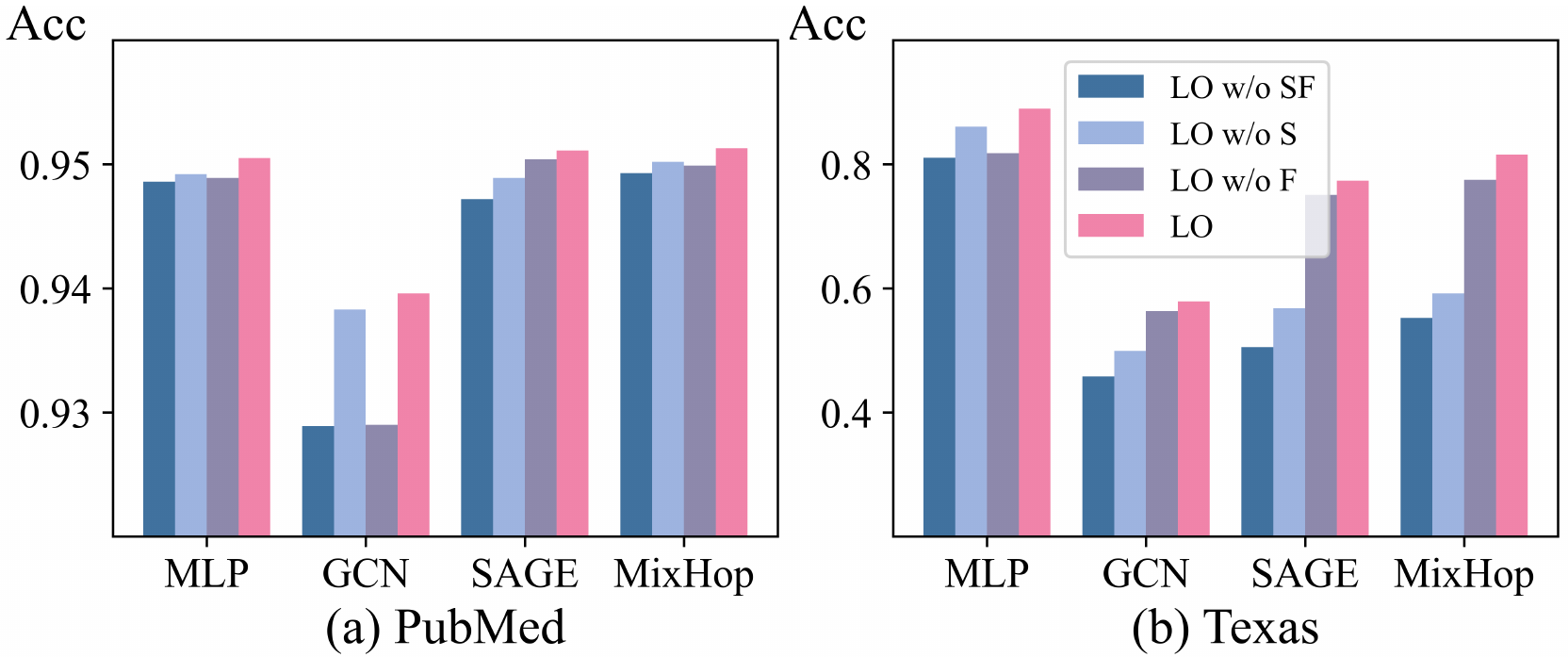}
  \caption{Ablation studies of L\scalebox{1}[0.8]{O}G\scalebox{1}[0.8]{IN} on PubMed and Texas. } 
  \label{fig:ablation}
\end{figure}

Regarding the structure refinement component,  L\scalebox{1}[0.8]{O}G\scalebox{1}[0.8]{IN} without edge pruning exhibits a significant performance decrease compared to the models with the complete coping mechanism, as shown in Fig.~\ref{fig:ablation}~(b). This observation aligns with our intuition, as in heterophilic graphs, the principal challenge for conventional GNNs in achieving generalization stems from the distinctiveness of their structural characteristics. 

By removing each key component separately, we verify that regardless of the dataset type, in this case, the extent of heterophily, our design of the fault-tolerant complementary analysis strategy for LLMs' responses is sound and necessary.

\subsection{Case Study~(RQ3)}
To answer the third research question, we select two individual nodes, respectively from Cora and Wisconsin, as specific examples to illustrate how L\scalebox{1}[0.8]{O}G\scalebox{1}[0.8]{IN} operates on them. These two nodes are both initially recognized as uncertain nodes and get misclassified by a pre-trained GNN. Through interaction with an LLM, the operation of feature enhancement or structure refinement is correspondingly conducted, thereby in turn helping the GNN make the right prediction.

\subsubsection{From a Citation Graph: Cora.}
\begin{figure*}
%\vspace{0cm} 
  \centering
  \includegraphics[width=0.95\linewidth]{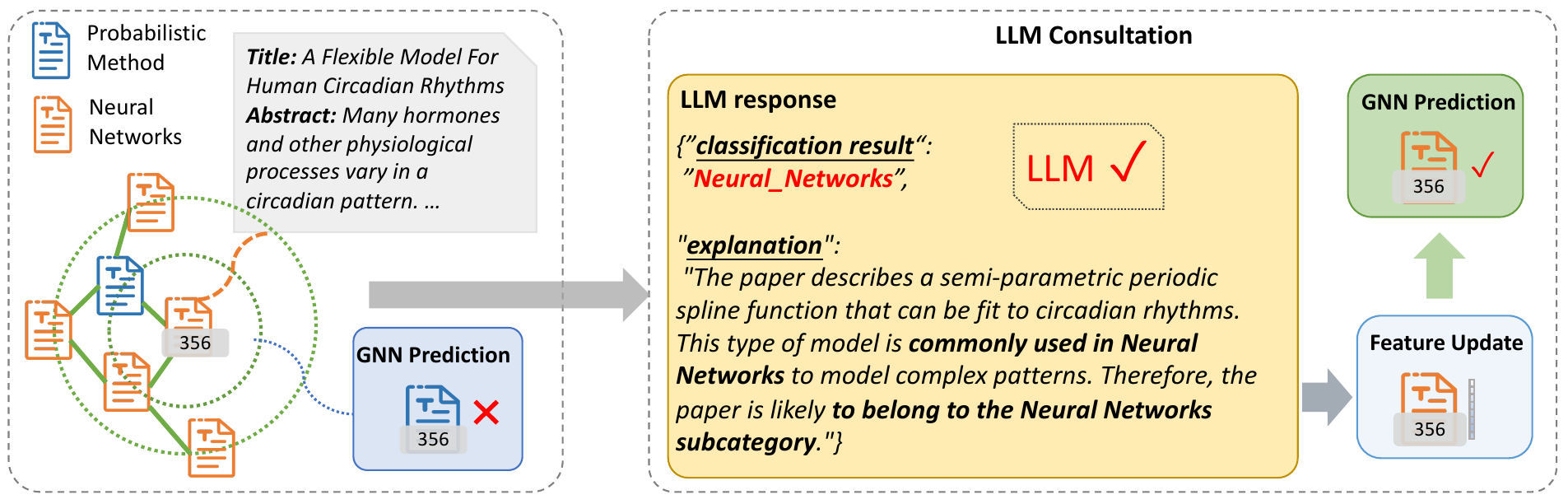}
  \caption{Case study of node 356 in Cora. Node 356 was first misclassified by GNN and selected as an uncertain node. The LLM consultation classified node 356 right along with a reasonable explanation. Then, the node representation was updated with the explanation. Finally, the retrained GNN managed to classify node 356 correctly.} 
  \label{fig:casecora}
\end{figure*}

We present node 356 from Cora as a representative example, whose ground truth label is \textit{Neural Networks}. Unlike other papers, the title and abstract of this paper do not feature its label as a term explicitly, which also poses challenges for human classification. Besides, node 356 only has two one-hop neighbors, one of which is labeled differently as \textit{Probabilistic Method}. This discrepancy may lead to failure in GNNs when processing this node.  Nevertheless, thanks to the comprehensive understanding of the paper content facilitated by the parametric knowledge of LLMs, accurate prediction and a concise rationale are generated. Consequently, the subsequent semantic enhancement of node features significantly contributes to the final prediction of the GNN. The consultation process is elaborated in Fig.~\ref{fig:casecora}.

\subsubsection{From a Web-page-link Graph: Wisconsin.}
\begin{figure*}
%\vspace{0.05cm} 
%\vspace{-0.05cm} 
  \centering
  \includegraphics[width=0.95\linewidth]{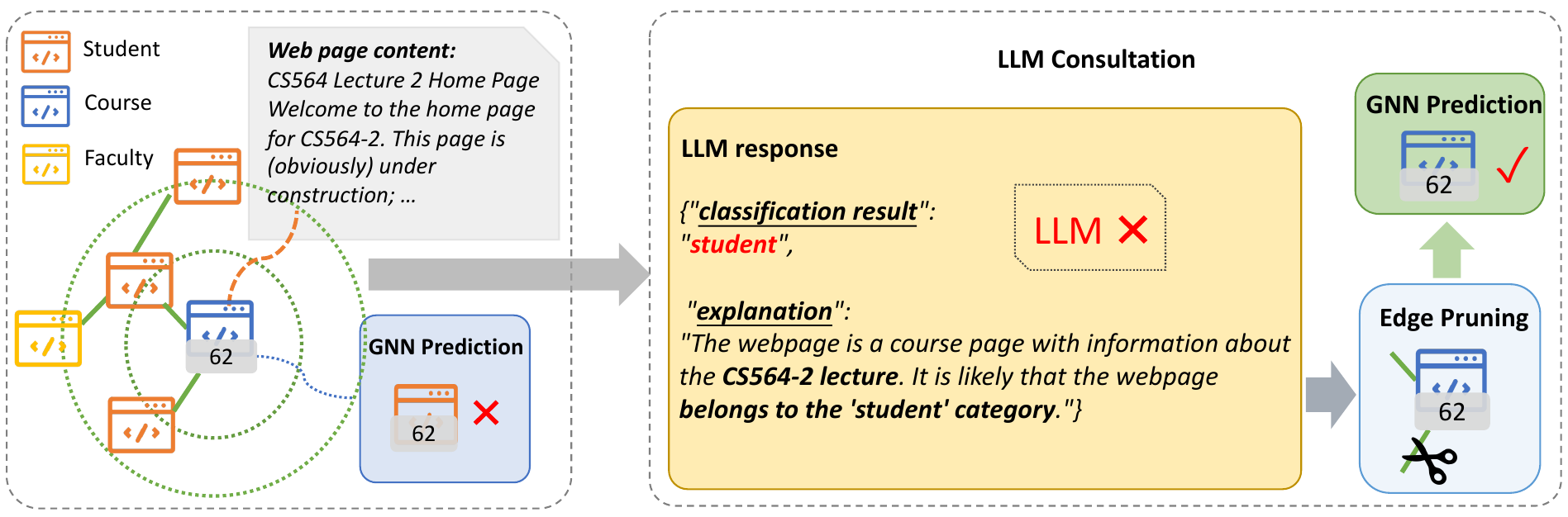}
  \caption{Case study of node 62 in Wisconsin. Node 62 was first misclassified by GNN and selected as an uncertain node. The LLM consultation classified node 62 wrong along with a self-contradictory explanation, which L\scalebox{1}[0.8]{O}G\scalebox{1}[0.8]{IN} left out. Then, edge pruning was performed around node 62. Finally, the retrained GNN managed to classify node 62 correctly.} 
  \label{fig:casewisconsin}
  %\vspace{0cm}
  %\setlength{\belowcaptionskip}{-0.2cm}
\end{figure*}

Node 62 from Wisconsin represents a web page of a course, with content that is clear enough for humans to identify as a \textit{course} homepage. However, due to its misleading neighborhood, where all nodes except itself in its 2-hop ego-graph do not represent course, among which 3 out of 4 are web pages of \textit{students}, pre-trained GNNs cannot directly classify node 8 accurately. Moreover, the LLM consultant also provides the incorrect classification, along with an illogical rationale mentioning GNN prediction as in Fig.~\ref{fig:casewisconsin}. This leads to edge pruning around node 62, which contributes to structure denoising that aids the re-trained GNN in making the right choice.

The studies of two interesting cases we encountered in experiments highlight the solidity and efficacy of L\scalebox{1}[0.8]{O}G\scalebox{1}[0.8]{IN} when handling various scenarios, which are consistent with our motivation of designing an LLM-fault-utilized strategy for feedback.

\subsection{Extensive Study~(RQ4)}
To answer the fourth question, we work on extended studies with more advanced LLMs, namely vicuna-v1.5-13b~\cite{zheng2023judging} and GPT 3.5-turbo-0125~\cite{brown2020language}. Due to the constraints related to computational resources and OpenAI API calling, we provide results solely for Cora, based on two runs as presented in Table~\ref{tab:extension}.

%\begin{figure}
  %\centering
  %\includegraphics[width=\linewidth]{figure/extension.png}
  %\caption{Performance Comparison among L\scalebox{1}[0.8]{O}G\scalebox{1}[0.8]{IN} with different LLMs on Cora. } 
  %\label{fig:extension}
%\end{figure}

In this experiment, all fundamental baselines with GPT 3.5-turbo-0125 outperform the ones with open-source LLMs. 
Additionally, the implementation of L\scalebox{1}[0.8]{O}G\scalebox{1}[0.8]{IN} with vicuna-v1.5-13b demonstrates a modest enhancement in predictive accuracy compared with vicuna-v1.5-7b, which has fewer parameters.
The trend in predictive accuracy aligns with our anticipations, indicating that the employment of a more advanced LLM within  L\scalebox{1}[0.8]{O}G\scalebox{1}[0.8]{IN} framework indeed facilitates performance increase. This underscores the significant potential of the LLMs-as-Consultants paradigm when equipped with more powerful LLMs.
\begin{table}[!htbp]
%\small
  \centering
    \caption{Performance comparison among L\scalebox{1}[0.8]{O}G\scalebox{1}[0.8]{IN} with different LLMs on Cora. }
\resizebox{\linewidth}{!}{
\begin{tabular}{cccc}
\toprule
LLMs & MLP & GCN & GraphSAGE \\ \midrule
Vicuna-v1.5-7b & 0.7063 $\pm$ 0.0331 & 0.8694 $\pm$ 0.0102 & 0.8727 $\pm$ 0.0201 \\
Vicuna-v1.5-13b & 0.7202 $\pm$ 0.0201 &  0.8702 $\pm$ 0.0191& 0.8739 $\pm$ 0.0135\\
GPT~3.5-turbo-0125 & 0.8123 $\pm$ 0.0254 & 0.8992 $\pm$ 0.0099 & 0.8856 $\pm$ 0.0102 \\ \bottomrule
\end{tabular}
}
\label{tab:extension}
\end{table}
\subsection{Comparison among LLM-based Paradigms~(RQ5)}
To answer the fifth research question, we investigate the LLMs-as-Predictors, LLMs-as-Enhancers, and our LLM-as-Consultants paradigms by testing typical methods derived from each. For LLMs-as-Predictors, we prompt vicuna-v1.5-7b~\cite{zheng2023judging} to directly get predictions. For LLMs-as-Enhancers, we adopt the TAPE method~\cite{he2023harnessing} equipped with llama2-13b-chat\cite{touvron2023llama}.

\begin{table*}[]

  \centering
    \caption{Performance comparison among LLMs-as-Predictors, LLMs-as-Enhancers and LLMs-as-Consultants paradigms.}
  \vspace{-2mm}
\begin{tabular}{ccccc}
\toprule
\multirow{2}{*}{Dataset} & \multirow{2}{*}{Method} & LLMs-as-Predictors & LLMs-as-Enhancers & LLMs-as-Consultants \\
                  &                   &(vicuna-v1.5-7b)  & (TAPE + llama2-13b-chat) & (L\scalebox{1}[0.8]{O}G\scalebox{1}[0.8]{IN} + vicuna-v1.5-7b)   \\ \midrule
\multirow{3}{*}{Cora} &     MLP         &  \multirow{3}{*}{0.7432 $\pm$ 0.0131}   &  0.7675 $\pm$ 0.0187   &0.7343 $\pm$ 0.0841 \\
                  &        GCN        &  & 0.8630 $\pm$ 0.0101 & \textbf{0.8759 $\pm$ 0.0151} \\
                  &       GraphSAGE       &     & 0.8625 $\pm$ 0.0093  & 0.8699 $\pm$ 0.0167 \\
                  \midrule
\multirow{3}{*}{PubMed} &      MLP         & \multirow{3}{*}{0.7847 $\pm$ 0.0632}    & 0.9475 $\pm$ 0.0046   & \textbf{0.9508 $\pm$ 0.0040}  \\
                  &          GCN         & & 0.9257 $\pm$ 0.0063  &  0.9401 $\pm$ 0.0043\\
                  &        GraphSAGE       &    & 0.9464 $\pm$ 0.0033  & 0.9505 $\pm$ 0.0031 \\
                  \midrule
\multirow{3}{*}{Arxiv-23} &   MLP          &  \multirow{3}{*}{0.7547 $\pm$ 0.0927}     & 0.7905 $\pm$ 0.0041 &  0.7909 $\pm$ 0.0064 \\
                  &          GCN      &   & 0.7751 $\pm$ 0.0029 & 0.7734 $\pm$ 0.0028   \\
                  &        GraphSAGE     &      &  0.7935 $\pm$ 0.0029   &\textbf{0.7961 $\pm$ 0.0029}  \\ \bottomrule
\end{tabular}
 \label{tab:exp:paradigms}%
\end{table*}

Note that the accuracy scores in Table~\ref{tab:exp:paradigms} differ from those in Table~\ref{tab:exp:main}, since we take the results of TAPE + llama2-13b-chat concerning their paper, which are reported for four runs. In Table~\ref{tab:exp:paradigms}, we also display the results of the same four data splits\footnote{\url{https://github.com/XiaoxinHe/TAPE}} to guarantee the comparison is fair. 

As Table~\ref{tab:exp:paradigms} shows, the LLMs-as-Consultants paradigm consistently outperforms the LLMs-as-Predictors paradigm across all datasets. Additionally, compared to the LLMs-as-Enhancers paradigm, our method surpasses the TAPE method equipped with llama2-13b-chat with only one exception, despite it employs an open-source LLM with significantly more parameters and prompts it with all nodes rather than a small subset selected. Our LLMs-as-Consultants paradigm demonstrates greater compatibility with lower time and resource consumption.

\subsection{Consulting with More Nodes~(RQ6)}
To address the sixth research question, we conducted experiments on the Cora and Wisconsin datasets using MLP and GraphSAGE to investigate the impact of consulting LLMs with more nodes. Our findings demonstrate that incorporating more nodes in the consultation process with LLMs can lead to significant performance improvements. The results are in Table~\ref{table:more_nodes}.

Through these experiments, we increased the number of nodes involved in the LLM consultation by 10\% each. The results on two example homophilic and heterophilic graphs show that as the number of consulted nodes increases, there are corresponding improvements in the overall performance of the model. This suggests that LLMs, when provided with a broader context through more extensive node consultation, can offer more insightful responses, thereby refining the model’s predictions and decisions. These findings underscore the potential benefits of leveraging LLMs more comprehensively, highlighting the effectiveness of the node consultation process to fully exploit the capabilities of LLMs in improving GNN performance.

\begin{table}[!htbp]
%\small
  \centering
    \caption{Performance comparison of consulting with more nodes on Cora and Wisconsin. }
\resizebox{\linewidth}{!}{
\begin{tabular}{ccc}
\toprule
Method & Cora & Wisconsin \\ \midrule
MLP+LOGIN & 0.7063 ± 0.0331 & 0.8528 ± 0.0588   \\
\textbf{MLP+LOGIN+10\% nodes} & 0.7147 ± 0.0318   &  0.8605 ± 0.0493      \\
GrraphSAGE+LOGIN & 0.8727 ± 0.0219  & 0.7434 ± 0.0930    \\ 
\textbf{GraphSAGE+LOGIN+10\% nodes} & 0.8789 ± 0.0310       & 0.7527 ± 0.0698    \\ 
\bottomrule
\end{tabular}
}
\label{table:more_nodes}
\end{table}

\subsection{L\scalebox{1}[0.8]{O}G\scalebox{1}[0.8]{IN} with Advanced GNNs~(RQ7)}
To address the seventh research question, we performed experiments incorporating APPNP and GCNII with L\scalebox{1}[0.8]{O}G\scalebox{1}[0.8]{IN} to demonstrate that L\scalebox{1}[0.8]{O}G\scalebox{1}[0.8]{IN} can also enhance the performance of advanced GNNs.

 The results in Table~\ref{table:advgnn} indicate notable performance gains on both Cora and Wisconsin. These enhancements underscore the versatility and broad applicability of L\scalebox{1}[0.8]{O}G\scalebox{1}[0.8]{IN} in enhancing advanced GNNs as well, reinforcing its potential as a general solution for boosting various types of GNNs. These experiments provide evidence that L\scalebox{1}[0.8]{O}G\scalebox{1}[0.8]{IN} is generally effective and not limited to improving specific types of GNNs.

\begin{table}[H]
%\small
  \centering
    \caption{Performance comparison of APPNP and GCNII trained with or without the L\scalebox{1}[0.8]{O}G\scalebox{1}[0.8]{IN} framework. }
\resizebox{\linewidth}{!}{
\begin{tabular}{ccc}
\toprule
Method & Cora & Wisconsin \\ \midrule
APPNP & 0.8539 ± 0.0477 & 0.7969 ± 0.0143    \\
APPNP+LOGIN & 0.8614 ± 0.0332 & 0.7973 ± 0.0202       \\
GCNII & 0.8833 ± 0.0027 & 0.7847 ± 0.0068     \\ 
GCNII+LOGIN & 0.8879 ± 0.0105 & 0.7903 ± 0.0121      \\ 
\bottomrule
\end{tabular}
}
\label{table:advgnn}
\end{table}

\section{Conclusion and Future Work}\label{sec:conclusion}
In this work, we propose a new paradigm of leveraging LLMs for graph tasks, coined ``LLMs-as-Consultants''. Following this paradigm, our L\scalebox{1}[0.8]{O}G\scalebox{1}[0.8]{IN} framework empowers interactive LLM consultation in the GNN training process. We identify uncertain nodes in the GNN pre-training stage, prompt LLMs with rich semantic and topological information compression, and parse LLMs' responses in a not only fault-tolerant but also fault-utilized way to enhance GNN re-training.
Extensive experiments on both homophilic and heterophilic graphs illustrate the validity and versatility of our proposed LLMs-as-Consultants paradigm.

For future endeavors, we intend to further refine GNN weights during the training phase using insights from LLMs' responses, building upon our current research. Additionally, how to efficiently involve a large scale of nodes in LLM consultation to improve the scalability of the LLMs-as-Consultants paradigm remains an open question.

%For future endeavors, we intend to further refine the weights of GNNs during the training phase by leveraging insights derived from the responses of LLMs, building upon our current research. By integrating the understanding and contextual knowledge provided by LLMs, we aim to enhance the accuracy and performance of GNNs in various applications.

Moreover, a significant challenge that remains is determining how to efficiently involve large-scale nodes in the LLM consultation process. This is crucial for improving the scalability and practicality of the LLMs-as-Consultants paradigm. Addressing this issue involves developing methods that can handle the extensive computational requirements and ensure that the consultation process remains effective and responsive even as the number of nodes increases. Exploring solutions such as distributed computing, optimized query strategies, and hierarchical consultation frameworks could be potential pathways to making this paradigm scalable and efficient for real-world applications.
\appendix
\appendixpage
\section{Example Prompts}
\label{sec:app:prompts}
We display 2 example node prompts on datasets Cora and Wisconsin respectively in Table~\ref{table:prompts} for reproducibility.
\begin{table*}[!htbp]

\caption{Example Prompts for Cora and Wisconsin.}
\centering
\begin{tabularx}{\textwidth}{lXXX}
\toprule
  \textbf{Datasets} & \textbf{Prompts} \\ 
  \midrule
Cora & Given a citation graph: \textit{\{ ``id'':``cora\_356'', ``graph'': \{ ``node\_idx'': 356, ``node\_list'': [356, 190, 510, 519, 2223, 192], ``one\_hop\_neighbors'': [190, 510], ``two\_hops\_neighbors'': [192, 519, 2223], ``node\_label'': [``Neural\_Networks'', ``Neural\_Networks'', ``Probabilistic\_Methods'', ``Neural\_Networks'', ``Neural\_Networks'', ``Neural\_Networks''] \}\}}, where the 0th node is the target paper, with the following information: Title: \textbf{[Title Text]}, Abstract: \textbf{[Abstract Text]}. And in the `node label' list / `GNN-predicted node label' list, you'll find the human-annotated / GNN-predicted subcategories corresponding to the neighbors within two hops of the target paper as per the `node\_list'. 
\newline\textbf{Question}: Which CS sub-category does this paper belong to? Give the most likely CS sub-categories of this paper directly, choosing from "Case\_Based", "Genetic\_Algorithms", "Neural\_Networks", "Probabilistic\_Methods", "Reinforcement\_Learning", "Rule\_Learning", "Theory". Ensure that your response can be parsed by Python json, using the following format as an example: \{"classification result": "Genetic\_Algorithms", "explanation": "your explanation for your classification here"\}. Ensure that the classification result must match one of the given choices.  \\ \midrule
Wisconsin & Given a webpage link graph: \textit{\{ ``id'': ``wisconsin\_62'', ``graph'': \{ ``node\_idx'': 62, ``node\_list'': [62, 166, 189, 165, 84], ``one\_hop\_neighbors'': [166, 189], ``two\_hops\_neighbors'': [84, 165], ``node\_label'': [``course'', ``student'', ``student'', ``student'', ``faculty''] \}\}}, where the 0th node is the target webpage, with the following content: \textbf{[webpage content text]}. And in the `node label' list / `GNN-predicted node label' list, you'll find the human-annotated / GNN-predicted subcategories corresponding to the neighbors within two hops of the target paper as per the `node\_list'. 
\newline\textbf{Question}: Which category does this webpage belong to? Give the most likely category of this webpage directly, choosing from ``course'', ``faculty'', ``student'', ``project'', ``staff''. Ensure that your response can be parsed by Python json, using the following format as an example: \{``classification result'': ``student'', ``explanation'': ``your explanation for your classification here'',\}. Ensure that the classification result must match one of the given choices. \\ \bottomrule
\end{tabularx}
%\captionsetup{labelfont={color=blue}}
\label{table:prompts}
\end{table*}

\section{more baselines of different LLMs-for-Graphs Paradigms}
To further show the effectiveness of the LLMs-as-Consultants paradigm, we conducted more experiments on Arxiv~\cite{lim2021large} dataset with more baselines of both LLMs-as-Predictors and LLMs-as-Enhancers paradigms as Table~\ref{table:exp:para_arxiv} shows. Note that Arxiv is a semi-homophilic graph with 169,343 nodes and 31,166,24 edges, among which 37\% nodes are heterophilic.

GraphGPT~\cite{tang2023graphgpt} and InstructGLM~\cite{ye2023natural} are LLMs-as-Predictors methods. The results of GraphGPT and InstructGLM on Arxiv dataset are directly borrowed from their papers.

SimTeG~\cite{duan2023simteg} and TAPE~\cite{he2023harnessing} are both LLMs-as-Enhancers methods. SimTeG involved LLM fine-tuning and TAPE prompted LLM for predictions and explanations to enhance the node features. For SimTeG, we trained GNNs with the released embeddings of Arxiv. For TAPE, we reproduced the prompting with Vicuna-7b-v1.5 to make the comparison fair. Note that SimTeG finetuned LLMs on all the training nodes and TAPE enhanced every node embedding, while our method only consults LLMs with some uncertain nodes, requiring less time and memory, but still performs on par with the baselines.

\begin{table}[htb]
\caption{Performance comparison with more baselines of LLMs-as-Predictors and LLMs-as-Enhancers paradigm on Arxiv.}
\centering
%\begin{tabularx}{\textwidth}{lXXX}
\begin{tabular}{ccc}
\textbf{Paradigm}                       & \textbf{Method}       & \textbf{Acc}             \\ \hline
LLMs-as-Predictors                   & GraphGPT              & 0.7511                   \\
LLMs-as-Predictors                   & InstructGLM           & 0.7570 ± 0.0012          \\ \hline
LLMs-as-Enhancers                    & MLP+SimTeG            & 0.7432 ± 0.0023          \\
LLMs-as-Enhancers                    & MLP+TAPE              & 0.7461 ± 0.0029          \\
\textbf{LLMs-as-Consultants}         & \textbf{MLP+LOGIN}    & \textbf{0.7476 ± 0.0036} \\ \hline
LLMs-as-Enhancers                    & GraphSAGE+SimTeG           & 0.7624 ± 0.0050          \\
LLMs-as-Enhancers                    & GraphSAGE+TAPE             & 0.7558± 0.0029           \\
\textbf{LLMs-as-Consultants}         & \textbf{GraphSAGE+LOGIN}   & \textbf{0.7609± 0.0041}  \\ \hline
\end{tabular}

\label{table:exp:para_arxiv}
\end{table}

%\begin{acks}
 %This work was supported by the [...] Research Fund of [...] (Number [...]). Additional funding was provided by [...] and [...]. We also thank [...] for contributing [...].
%\end{acks}

%\clearpage
\normalem
\bibliographystyle{ACM-Reference-Format}
\bibliography{sample}


\begin{thebibliography}{96}

%%% the Web address.

\ifx \showCODEN    \undefined \def \showCODEN     #1{\unskip}     \fi
\ifx \showDOI      \undefined \def \showDOI       #1{#1}\fi
\ifx \showISBNx    \undefined \def \showISBNx     #1{\unskip}     \fi
\ifx \showISBNxiii \undefined \def \showISBNxiii  #1{\unskip}     \fi
\ifx \showISSN     \undefined \def \showISSN      #1{\unskip}     \fi
\ifx \showLCCN     \undefined \def \showLCCN      #1{\unskip}     \fi
\ifx \shownote     \undefined \def \shownote      #1{#1}          \fi
\ifx \showarticletitle \undefined \def \showarticletitle #1{#1}   \fi
\ifx \showURL      \undefined \def \showURL       {\relax}        \fi
% invisible to TeX
\providecommand\bibfield[2]{#2}
\providecommand\bibinfo[2]{#2}
\providecommand\natexlab[1]{#1}
\providecommand\showeprint[2][]{arXiv:#2}

\bibitem[\protect\citeauthoryear{Abu-El-Haija, Perozzi, Kapoor, Alipourfard, Lerman, Harutyunyan, Ver~Steeg, and Galstyan}{Abu-El-Haija et~al\mbox{.}}{2019}]%
        {abu2019mixhop}
\bibfield{author}{\bibinfo{person}{Sami Abu-El-Haija}, \bibinfo{person}{Bryan Perozzi}, \bibinfo{person}{Amol Kapoor}, \bibinfo{person}{Nazanin Alipourfard}, \bibinfo{person}{Kristina Lerman}, \bibinfo{person}{Hrayr Harutyunyan}, \bibinfo{person}{Greg Ver~Steeg}, {and} \bibinfo{person}{Aram Galstyan}.} \bibinfo{year}{2019}\natexlab{}.
\newblock \showarticletitle{Mixhop: Higher-order graph convolutional architectures via sparsified neighborhood mixing}. In \bibinfo{booktitle}{\emph{international conference on machine learning}}. PMLR, \bibinfo{pages}{21--29}.
\newblock


\bibitem[\protect\citeauthoryear{Akoglu, Tong, and Koutra}{Akoglu et~al\mbox{.}}{2014}]%
        {akoglu2014graphbased}
\bibfield{author}{\bibinfo{person}{Leman Akoglu}, \bibinfo{person}{Hanghang Tong}, {and} \bibinfo{person}{Danai Koutra}.} \bibinfo{year}{2014}\natexlab{}.
\newblock \bibinfo{title}{Graph-based Anomaly Detection and Description: A Survey}.
\newblock
\newblock
\showeprint[arxiv]{1404.4679}~[cs.SI]


\bibitem[\protect\citeauthoryear{Alayrac, Donahue, Luc, Miech, Barr, Hasson, Lenc, Mensch, Millican, Reynolds, et~al\mbox{.}}{Alayrac et~al\mbox{.}}{2022}]%
        {alayrac2022flamingo}
\bibfield{author}{\bibinfo{person}{Jean-Baptiste Alayrac}, \bibinfo{person}{Jeff Donahue}, \bibinfo{person}{Pauline Luc}, \bibinfo{person}{Antoine Miech}, \bibinfo{person}{Iain Barr}, \bibinfo{person}{Yana Hasson}, \bibinfo{person}{Karel Lenc}, \bibinfo{person}{Arthur Mensch}, \bibinfo{person}{Katherine Millican}, \bibinfo{person}{Malcolm Reynolds}, {et~al\mbox{.}}} \bibinfo{year}{2022}\natexlab{}.
\newblock \showarticletitle{Flamingo: a visual language model for few-shot learning}.
\newblock \bibinfo{journal}{\emph{Advances in Neural Information Processing Systems}}  \bibinfo{volume}{35} (\bibinfo{year}{2022}), \bibinfo{pages}{23716--23736}.
\newblock


\bibitem[\protect\citeauthoryear{Alcacer and Gittelman}{Alcacer and Gittelman}{2006}]%
        {alcacer2006patent}
\bibfield{author}{\bibinfo{person}{Juan Alcacer} {and} \bibinfo{person}{Michelle Gittelman}.} \bibinfo{year}{2006}\natexlab{}.
\newblock \showarticletitle{Patent citations as a measure of knowledge flows: The influence of examiner citations}.
\newblock \bibinfo{journal}{\emph{The review of economics and statistics}} \bibinfo{volume}{88}, \bibinfo{number}{4} (\bibinfo{year}{2006}), \bibinfo{pages}{774--779}.
\newblock


\bibitem[\protect\citeauthoryear{Basu, Hirsh, and Cohen}{Basu et~al\mbox{.}}{1998}]%
        {1998Recommendation}
\bibfield{author}{\bibinfo{person}{Chumki Basu}, \bibinfo{person}{Haym Hirsh}, {and} \bibinfo{person}{William Cohen}.} \bibinfo{year}{1998}\natexlab{}.
\newblock \showarticletitle{Recommendation as Classification: Using Social and Content-Based Information in Recommendation}.
\newblock \bibinfo{journal}{\emph{Proceedings of AAAI-98}} (\bibinfo{year}{1998}).
\newblock


\bibitem[\protect\citeauthoryear{Battaglia, Hamrick, Bapst, Sanchez-Gonzalez, Zambaldi, Malinowski, Tacchetti, Raposo, Santoro, Faulkner, et~al\mbox{.}}{Battaglia et~al\mbox{.}}{2018}]%
        {battaglia2018relational}
\bibfield{author}{\bibinfo{person}{Peter~W Battaglia}, \bibinfo{person}{Jessica~B Hamrick}, \bibinfo{person}{Victor Bapst}, \bibinfo{person}{Alvaro Sanchez-Gonzalez}, \bibinfo{person}{Vinicius Zambaldi}, \bibinfo{person}{Mateusz Malinowski}, \bibinfo{person}{Andrea Tacchetti}, \bibinfo{person}{David Raposo}, \bibinfo{person}{Adam Santoro}, \bibinfo{person}{Ryan Faulkner}, {et~al\mbox{.}}} \bibinfo{year}{2018}\natexlab{}.
\newblock \showarticletitle{Relational inductive biases, deep learning, and graph networks}.
\newblock \bibinfo{journal}{\emph{arXiv preprint arXiv:1806.01261}} (\bibinfo{year}{2018}).
\newblock


\bibitem[\protect\citeauthoryear{Blundell}{Blundell}{1996}]%
        {blundell1996structure}
\bibfield{author}{\bibinfo{person}{Tom~L Blundell}.} \bibinfo{year}{1996}\natexlab{}.
\newblock \showarticletitle{Structure-based drug design}.
\newblock \bibinfo{journal}{\emph{Nature}} \bibinfo{volume}{384}, \bibinfo{number}{6604} (\bibinfo{year}{1996}), \bibinfo{pages}{23}.
\newblock


\bibitem[\protect\citeauthoryear{Bo, Wang, Shi, and Shen}{Bo et~al\mbox{.}}{2021}]%
        {bo2021beyond}
\bibfield{author}{\bibinfo{person}{Deyu Bo}, \bibinfo{person}{Xiao Wang}, \bibinfo{person}{Chuan Shi}, {and} \bibinfo{person}{Huawei Shen}.} \bibinfo{year}{2021}\natexlab{}.
\newblock \showarticletitle{Beyond low-frequency information in graph convolutional networks}. In \bibinfo{booktitle}{\emph{Proceedings of the AAAI Conference on Artificial Intelligence}}, Vol.~\bibinfo{volume}{35}. \bibinfo{pages}{3950--3957}.
\newblock


\bibitem[\protect\citeauthoryear{Broder, Andrei, Kumar, Ravi, and Janet}{Broder et~al\mbox{.}}{2000}]%
        {Broder2000Graph}
\bibfield{author}{\bibinfo{person}{Broder}, \bibinfo{person}{Andrei}, \bibinfo{person}{Kumar}, \bibinfo{person}{Ravi}, {and} \bibinfo{person}{Janet}.} \bibinfo{year}{2000}\natexlab{}.
\newblock \showarticletitle{Graph Structure in the Web : Experiments and models}.
\newblock  (\bibinfo{year}{2000}).
\newblock


\bibitem[\protect\citeauthoryear{Brown, Mann, Ryder, Subbiah, Kaplan, Dhariwal, Neelakantan, Shyam, Sastry, Askell, et~al\mbox{.}}{Brown et~al\mbox{.}}{2020}]%
        {brown2020language}
\bibfield{author}{\bibinfo{person}{Tom Brown}, \bibinfo{person}{Benjamin Mann}, \bibinfo{person}{Nick Ryder}, \bibinfo{person}{Melanie Subbiah}, \bibinfo{person}{Jared~D Kaplan}, \bibinfo{person}{Prafulla Dhariwal}, \bibinfo{person}{Arvind Neelakantan}, \bibinfo{person}{Pranav Shyam}, \bibinfo{person}{Girish Sastry}, \bibinfo{person}{Amanda Askell}, {et~al\mbox{.}}} \bibinfo{year}{2020}\natexlab{}.
\newblock \showarticletitle{Language models are few-shot learners}.
\newblock \bibinfo{journal}{\emph{Advances in neural information processing systems}}  \bibinfo{volume}{33} (\bibinfo{year}{2020}), \bibinfo{pages}{1877--1901}.
\newblock


\bibitem[\protect\citeauthoryear{Chai, Zhang, Wu, Han, Hu, Huang, and Yang}{Chai et~al\mbox{.}}{2023}]%
        {chai2023graphllm}
\bibfield{author}{\bibinfo{person}{Ziwei Chai}, \bibinfo{person}{Tianjie Zhang}, \bibinfo{person}{Liang Wu}, \bibinfo{person}{Kaiqiao Han}, \bibinfo{person}{Xiaohai Hu}, \bibinfo{person}{Xuanwen Huang}, {and} \bibinfo{person}{Yang Yang}.} \bibinfo{year}{2023}\natexlab{}.
\newblock \showarticletitle{Graphllm: Boosting graph reasoning ability of large language model}.
\newblock \bibinfo{journal}{\emph{arXiv preprint arXiv:2310.05845}} (\bibinfo{year}{2023}).
\newblock


\bibitem[\protect\citeauthoryear{Chen, Lin, Zhao, Ren, Li, Zhou, and Sun}{Chen et~al\mbox{.}}{2021}]%
        {chen2021topology}
\bibfield{author}{\bibinfo{person}{Deli Chen}, \bibinfo{person}{Yankai Lin}, \bibinfo{person}{Guangxiang Zhao}, \bibinfo{person}{Xuancheng Ren}, \bibinfo{person}{Peng Li}, \bibinfo{person}{Jie Zhou}, {and} \bibinfo{person}{Xu Sun}.} \bibinfo{year}{2021}\natexlab{}.
\newblock \showarticletitle{Topology-imbalance learning for semi-supervised node classification}.
\newblock \bibinfo{journal}{\emph{Advances in Neural Information Processing Systems}}  \bibinfo{volume}{34} (\bibinfo{year}{2021}), \bibinfo{pages}{29885--29897}.
\newblock


\bibitem[\protect\citeauthoryear{Chen, Wei, Huang, Ding, and Li}{Chen et~al\mbox{.}}{2020}]%
        {chen2020simple}
\bibfield{author}{\bibinfo{person}{Ming Chen}, \bibinfo{person}{Zhewei Wei}, \bibinfo{person}{Zengfeng Huang}, \bibinfo{person}{Bolin Ding}, {and} \bibinfo{person}{Yaliang Li}.} \bibinfo{year}{2020}\natexlab{}.
\newblock \showarticletitle{Simple and deep graph convolutional networks}. In \bibinfo{booktitle}{\emph{International conference on machine learning}}. PMLR, \bibinfo{pages}{1725--1735}.
\newblock


\bibitem[\protect\citeauthoryear{Chen, Mao, Li, Jin, Wen, Wei, Wang, Yin, Fan, Liu, et~al\mbox{.}}{Chen et~al\mbox{.}}{2023}]%
        {chen2023exploring}
\bibfield{author}{\bibinfo{person}{Zhikai Chen}, \bibinfo{person}{Haitao Mao}, \bibinfo{person}{Hang Li}, \bibinfo{person}{Wei Jin}, \bibinfo{person}{Hongzhi Wen}, \bibinfo{person}{Xiaochi Wei}, \bibinfo{person}{Shuaiqiang Wang}, \bibinfo{person}{Dawei Yin}, \bibinfo{person}{Wenqi Fan}, \bibinfo{person}{Hui Liu}, {et~al\mbox{.}}} \bibinfo{year}{2023}\natexlab{}.
\newblock \showarticletitle{Exploring the potential of large language models (llms) in learning on graphs}.
\newblock \bibinfo{journal}{\emph{arXiv preprint arXiv:2307.03393}} (\bibinfo{year}{2023}).
\newblock


\bibitem[\protect\citeauthoryear{Craven, DiPasquo, Freitag, McCallum, Mitchell, Nigam, and Slattery}{Craven et~al\mbox{.}}{1998}]%
        {craven1998learning}
\bibfield{author}{\bibinfo{person}{Mark Craven}, \bibinfo{person}{Dan DiPasquo}, \bibinfo{person}{Dayne Freitag}, \bibinfo{person}{Andrew McCallum}, \bibinfo{person}{Tom Mitchell}, \bibinfo{person}{Kamal Nigam}, {and} \bibinfo{person}{Se{\'a}n Slattery}.} \bibinfo{year}{1998}\natexlab{}.
\newblock \showarticletitle{Learning to extract symbolic knowledge from the World Wide Web}.
\newblock \bibinfo{journal}{\emph{AAAI/IAAI}} \bibinfo{volume}{3}, \bibinfo{number}{3.6} (\bibinfo{year}{1998}), \bibinfo{pages}{2}.
\newblock


\bibitem[\protect\citeauthoryear{Devlin, Chang, Lee, and Toutanova}{Devlin et~al\mbox{.}}{2018}]%
        {devlin2018bert}
\bibfield{author}{\bibinfo{person}{Jacob Devlin}, \bibinfo{person}{Ming-Wei Chang}, \bibinfo{person}{Kenton Lee}, {and} \bibinfo{person}{Kristina Toutanova}.} \bibinfo{year}{2018}\natexlab{}.
\newblock \showarticletitle{Bert: Pre-training of deep bidirectional transformers for language understanding}.
\newblock \bibinfo{journal}{\emph{arXiv preprint arXiv:1810.04805}} (\bibinfo{year}{2018}).
\newblock


\bibitem[\protect\citeauthoryear{Duan, Liu, Chua, Yan, Ooi, Xie, and He}{Duan et~al\mbox{.}}{2023}]%
        {duan2023simteg}
\bibfield{author}{\bibinfo{person}{Keyu Duan}, \bibinfo{person}{Qian Liu}, \bibinfo{person}{Tat-Seng Chua}, \bibinfo{person}{Shuicheng Yan}, \bibinfo{person}{Wei~Tsang Ooi}, \bibinfo{person}{Qizhe Xie}, {and} \bibinfo{person}{Junxian He}.} \bibinfo{year}{2023}\natexlab{}.
\newblock \showarticletitle{Simteg: A frustratingly simple approach improves textual graph learning}.
\newblock \bibinfo{journal}{\emph{arXiv preprint arXiv:2308.02565}} (\bibinfo{year}{2023}).
\newblock


\bibitem[\protect\citeauthoryear{Fatemi, Halcrow, and Perozzi}{Fatemi et~al\mbox{.}}{2023}]%
        {fatemi2023talk}
\bibfield{author}{\bibinfo{person}{Bahare Fatemi}, \bibinfo{person}{Jonathan Halcrow}, {and} \bibinfo{person}{Bryan Perozzi}.} \bibinfo{year}{2023}\natexlab{}.
\newblock \showarticletitle{Talk like a graph: Encoding graphs for large language models}.
\newblock \bibinfo{journal}{\emph{arXiv preprint arXiv:2310.04560}} (\bibinfo{year}{2023}).
\newblock


\bibitem[\protect\citeauthoryear{Fu, Zhang, Meng, and King}{Fu et~al\mbox{.}}{2020}]%
        {fu2020magnn}
\bibfield{author}{\bibinfo{person}{Xinyu Fu}, \bibinfo{person}{Jiani Zhang}, \bibinfo{person}{Ziqiao Meng}, {and} \bibinfo{person}{Irwin King}.} \bibinfo{year}{2020}\natexlab{}.
\newblock \showarticletitle{Magnn: Metapath aggregated graph neural network for heterogeneous graph embedding}. In \bibinfo{booktitle}{\emph{Proceedings of the web conference 2020}}. \bibinfo{pages}{2331--2341}.
\newblock


\bibitem[\protect\citeauthoryear{Gasteiger, Bojchevski, and G{\"u}nnemann}{Gasteiger et~al\mbox{.}}{2018}]%
        {gasteiger2018predict}
\bibfield{author}{\bibinfo{person}{Johannes Gasteiger}, \bibinfo{person}{Aleksandar Bojchevski}, {and} \bibinfo{person}{Stephan G{\"u}nnemann}.} \bibinfo{year}{2018}\natexlab{}.
\newblock \showarticletitle{Predict then propagate: Graph neural networks meet personalized pagerank}.
\newblock \bibinfo{journal}{\emph{arXiv preprint arXiv:1810.05997}} (\bibinfo{year}{2018}).
\newblock


\bibitem[\protect\citeauthoryear{Gilmer, Schoenholz, Riley, Vinyals, and Dahl}{Gilmer et~al\mbox{.}}{2017}]%
        {gilmer2017neural}
\bibfield{author}{\bibinfo{person}{Justin Gilmer}, \bibinfo{person}{Samuel~S Schoenholz}, \bibinfo{person}{Patrick~F Riley}, \bibinfo{person}{Oriol Vinyals}, {and} \bibinfo{person}{George~E Dahl}.} \bibinfo{year}{2017}\natexlab{}.
\newblock \showarticletitle{Neural message passing for quantum chemistry}. In \bibinfo{booktitle}{\emph{International conference on machine learning}}. PMLR, \bibinfo{pages}{1263--1272}.
\newblock


\bibitem[\protect\citeauthoryear{Golbeck and Hendler}{Golbeck and Hendler}{2006}]%
        {2006Inferring}
\bibfield{author}{\bibinfo{person}{J. Golbeck} {and} \bibinfo{person}{J Hendler}.} \bibinfo{year}{2006}\natexlab{}.
\newblock \showarticletitle{Inferring Trust Relationships in Web-Based Social Networks}.
\newblock \bibinfo{journal}{\emph{Acm Transactions on Internet Technology}} (\bibinfo{year}{2006}).
\newblock


\bibitem[\protect\citeauthoryear{Guo, Du, and Liu}{Guo et~al\mbox{.}}{2023}]%
        {guo2023gpt4graph}
\bibfield{author}{\bibinfo{person}{Jiayan Guo}, \bibinfo{person}{Lun Du}, {and} \bibinfo{person}{Hengyu Liu}.} \bibinfo{year}{2023}\natexlab{}.
\newblock \showarticletitle{GPT4Graph: Can Large Language Models Understand Graph Structured Data? An Empirical Evaluation and Benchmarking}.
\newblock \bibinfo{journal}{\emph{arXiv preprint arXiv:2305.15066}} (\bibinfo{year}{2023}).
\newblock


\bibitem[\protect\citeauthoryear{Hamilton, Ying, and Leskovec}{Hamilton et~al\mbox{.}}{2017a}]%
        {hamilton2017inductive}
\bibfield{author}{\bibinfo{person}{Will Hamilton}, \bibinfo{person}{Zhitao Ying}, {and} \bibinfo{person}{Jure Leskovec}.} \bibinfo{year}{2017}\natexlab{a}.
\newblock \showarticletitle{Inductive representation learning on large graphs}.
\newblock \bibinfo{journal}{\emph{Advances in neural information processing systems}}  \bibinfo{volume}{30} (\bibinfo{year}{2017}).
\newblock


\bibitem[\protect\citeauthoryear{Hamilton}{Hamilton}{2020}]%
        {hamilton2020graph}
\bibfield{author}{\bibinfo{person}{William~L Hamilton}.} \bibinfo{year}{2020}\natexlab{}.
\newblock \bibinfo{booktitle}{\emph{Graph representation learning}}.
\newblock \bibinfo{publisher}{Morgan \& Claypool Publishers}.
\newblock


\bibitem[\protect\citeauthoryear{Hamilton, Ying, and Leskovec}{Hamilton et~al\mbox{.}}{2017b}]%
        {hamilton2017representation}
\bibfield{author}{\bibinfo{person}{William~L Hamilton}, \bibinfo{person}{Rex Ying}, {and} \bibinfo{person}{Jure Leskovec}.} \bibinfo{year}{2017}\natexlab{b}.
\newblock \showarticletitle{Representation learning on graphs: Methods and applications}.
\newblock \bibinfo{journal}{\emph{arXiv preprint arXiv:1709.05584}} (\bibinfo{year}{2017}).
\newblock


\bibitem[\protect\citeauthoryear{He, Liu, Gao, and Chen}{He et~al\mbox{.}}{2021}]%
        {he2021deberta}
\bibfield{author}{\bibinfo{person}{Pengcheng He}, \bibinfo{person}{Xiaodong Liu}, \bibinfo{person}{Jianfeng Gao}, {and} \bibinfo{person}{Weizhu Chen}.} \bibinfo{year}{2021}\natexlab{}.
\newblock \showarticletitle{DEBERTA: DECODING-ENHANCED BERT WITH DISENTANGLED ATTENTION}. In \bibinfo{booktitle}{\emph{International Conference on Learning Representations}}.
\newblock
\urldef\tempurl%
\url{https://openreview.net/forum?id=XPZIaotutsD}
\showURL{%
\tempurl}


\bibitem[\protect\citeauthoryear{He, Bresson, Laurent, Perold, LeCun, and Hooi}{He et~al\mbox{.}}{2023}]%
        {he2023harnessing}
\bibfield{author}{\bibinfo{person}{Xiaoxin He}, \bibinfo{person}{Xavier Bresson}, \bibinfo{person}{Thomas Laurent}, \bibinfo{person}{Adam Perold}, \bibinfo{person}{Yann LeCun}, {and} \bibinfo{person}{Bryan Hooi}.} \bibinfo{year}{2023}\natexlab{}.
\newblock \showarticletitle{Harnessing Explanations: LLM-to-LM Interpreter for Enhanced Text-Attributed Graph Representation Learning}.
\newblock \bibinfo{journal}{\emph{arXiv preprint arXiv:2305.19523}} (\bibinfo{year}{2023}).
\newblock


\bibitem[\protect\citeauthoryear{Hou, Ye, Song, and Abdulhayoglu}{Hou et~al\mbox{.}}{2017}]%
        {hou2017hindroid}
\bibfield{author}{\bibinfo{person}{Shifu Hou}, \bibinfo{person}{Yanfang Ye}, \bibinfo{person}{Yangqiu Song}, {and} \bibinfo{person}{Melih Abdulhayoglu}.} \bibinfo{year}{2017}\natexlab{}.
\newblock \showarticletitle{Hindroid: An intelligent android malware detection system based on structured heterogeneous information network}. In \bibinfo{booktitle}{\emph{Proceedings of the 23rd ACM SIGKDD international conference on knowledge discovery and data mining}}. \bibinfo{pages}{1507--1515}.
\newblock


\bibitem[\protect\citeauthoryear{Hu, Shi, Zhao, and Yu}{Hu et~al\mbox{.}}{2018}]%
        {hu2018leveraging}
\bibfield{author}{\bibinfo{person}{Binbin Hu}, \bibinfo{person}{Chuan Shi}, \bibinfo{person}{Wayne~Xin Zhao}, {and} \bibinfo{person}{Philip~S Yu}.} \bibinfo{year}{2018}\natexlab{}.
\newblock \showarticletitle{Leveraging meta-path based context for top-n recommendation with a neural co-attention model}. In \bibinfo{booktitle}{\emph{Proceedings of the 24th ACM SIGKDD international conference on knowledge discovery \& data mining}}. \bibinfo{pages}{1531--1540}.
\newblock


\bibitem[\protect\citeauthoryear{Hu, Zhang, Shi, Zhou, Li, and Qi}{Hu et~al\mbox{.}}{2019}]%
        {hu2019cash}
\bibfield{author}{\bibinfo{person}{Binbin Hu}, \bibinfo{person}{Zhiqiang Zhang}, \bibinfo{person}{Chuan Shi}, \bibinfo{person}{Jun Zhou}, \bibinfo{person}{Xiaolong Li}, {and} \bibinfo{person}{Yuan Qi}.} \bibinfo{year}{2019}\natexlab{}.
\newblock \showarticletitle{Cash-out user detection based on attributed heterogeneous information network with a hierarchical attention mechanism}. In \bibinfo{booktitle}{\emph{Proceedings of the AAAI Conference on Artificial Intelligence}}, Vol.~\bibinfo{volume}{33}. \bibinfo{pages}{946--953}.
\newblock


\bibitem[\protect\citeauthoryear{Hu, Shen, Wallis, Allen-Zhu, Li, Wang, Wang, and Chen}{Hu et~al\mbox{.}}{2021}]%
        {hu2021lora}
\bibfield{author}{\bibinfo{person}{Edward~J Hu}, \bibinfo{person}{Yelong Shen}, \bibinfo{person}{Phillip Wallis}, \bibinfo{person}{Zeyuan Allen-Zhu}, \bibinfo{person}{Yuanzhi Li}, \bibinfo{person}{Shean Wang}, \bibinfo{person}{Lu Wang}, {and} \bibinfo{person}{Weizhu Chen}.} \bibinfo{year}{2021}\natexlab{}.
\newblock \showarticletitle{Lora: Low-rank adaptation of large language models}.
\newblock \bibinfo{journal}{\emph{arXiv preprint arXiv:2106.09685}} (\bibinfo{year}{2021}).
\newblock


\bibitem[\protect\citeauthoryear{Huang and Chang}{Huang and Chang}{2022}]%
        {huang2022towards}
\bibfield{author}{\bibinfo{person}{Jie Huang} {and} \bibinfo{person}{Kevin Chen-Chuan Chang}.} \bibinfo{year}{2022}\natexlab{}.
\newblock \showarticletitle{Towards reasoning in large language models: A survey}.
\newblock \bibinfo{journal}{\emph{arXiv preprint arXiv:2212.10403}} (\bibinfo{year}{2022}).
\newblock


\bibitem[\protect\citeauthoryear{Huang, Liu, Ao, Li, Chi, Feng, Yang, and He}{Huang et~al\mbox{.}}{2022}]%
        {huang2022auc}
\bibfield{author}{\bibinfo{person}{Mengda Huang}, \bibinfo{person}{Yang Liu}, \bibinfo{person}{Xiang Ao}, \bibinfo{person}{Kuan Li}, \bibinfo{person}{Jianfeng Chi}, \bibinfo{person}{Jinghua Feng}, \bibinfo{person}{Hao Yang}, {and} \bibinfo{person}{Qing He}.} \bibinfo{year}{2022}\natexlab{}.
\newblock \showarticletitle{Auc-oriented graph neural network for fraud detection}. In \bibinfo{booktitle}{\emph{Proceedings of the ACM web conference 2022}}. \bibinfo{pages}{1311--1321}.
\newblock


\bibitem[\protect\citeauthoryear{Izadi, Fang, Stevenson, and Lin}{Izadi et~al\mbox{.}}{2020}]%
        {izadi2020optimization}
\bibfield{author}{\bibinfo{person}{Mohammad~Rasool Izadi}, \bibinfo{person}{Yihao Fang}, \bibinfo{person}{Robert Stevenson}, {and} \bibinfo{person}{Lizhen Lin}.} \bibinfo{year}{2020}\natexlab{}.
\newblock \showarticletitle{Optimization of graph neural networks with natural gradient descent}. In \bibinfo{booktitle}{\emph{2020 IEEE international conference on big data (big data)}}. IEEE, \bibinfo{pages}{171--179}.
\newblock


\bibitem[\protect\citeauthoryear{Ju, Yi, Wang, Xiao, Mao, Li, Gu, Qin, Yin, Wang, et~al\mbox{.}}{Ju et~al\mbox{.}}{2024}]%
        {ju2024survey}
\bibfield{author}{\bibinfo{person}{Wei Ju}, \bibinfo{person}{Siyu Yi}, \bibinfo{person}{Yifan Wang}, \bibinfo{person}{Zhiping Xiao}, \bibinfo{person}{Zhengyang Mao}, \bibinfo{person}{Hourun Li}, \bibinfo{person}{Yiyang Gu}, \bibinfo{person}{Yifang Qin}, \bibinfo{person}{Nan Yin}, \bibinfo{person}{Senzhang Wang}, {et~al\mbox{.}}} \bibinfo{year}{2024}\natexlab{}.
\newblock \showarticletitle{A survey of graph neural networks in real world: Imbalance, noise, privacy and ood challenges}.
\newblock \bibinfo{journal}{\emph{arXiv preprint arXiv:2403.04468}} (\bibinfo{year}{2024}).
\newblock


\bibitem[\protect\citeauthoryear{Kaplan, McCandlish, Henighan, Brown, Chess, Child, Gray, Radford, Wu, and Amodei}{Kaplan et~al\mbox{.}}{2020}]%
        {kaplan2020scaling}
\bibfield{author}{\bibinfo{person}{Jared Kaplan}, \bibinfo{person}{Sam McCandlish}, \bibinfo{person}{Tom Henighan}, \bibinfo{person}{Tom~B Brown}, \bibinfo{person}{Benjamin Chess}, \bibinfo{person}{Rewon Child}, \bibinfo{person}{Scott Gray}, \bibinfo{person}{Alec Radford}, \bibinfo{person}{Jeffrey Wu}, {and} \bibinfo{person}{Dario Amodei}.} \bibinfo{year}{2020}\natexlab{}.
\newblock \showarticletitle{Scaling laws for neural language models}.
\newblock \bibinfo{journal}{\emph{arXiv preprint arXiv:2001.08361}} (\bibinfo{year}{2020}).
\newblock


\bibitem[\protect\citeauthoryear{Kearnes, McCloskey, Berndl, Pande, and Riley}{Kearnes et~al\mbox{.}}{2016}]%
        {kearnes2016molecular}
\bibfield{author}{\bibinfo{person}{Steven Kearnes}, \bibinfo{person}{Kevin McCloskey}, \bibinfo{person}{Marc Berndl}, \bibinfo{person}{Vijay Pande}, {and} \bibinfo{person}{Patrick Riley}.} \bibinfo{year}{2016}\natexlab{}.
\newblock \showarticletitle{Molecular graph convolutions: moving beyond fingerprints}.
\newblock \bibinfo{journal}{\emph{Journal of computer-aided molecular design}}  \bibinfo{volume}{30} (\bibinfo{year}{2016}), \bibinfo{pages}{595--608}.
\newblock


\bibitem[\protect\citeauthoryear{Kendall and Gal}{Kendall and Gal}{2017}]%
        {kendall2017uncertainties}
\bibfield{author}{\bibinfo{person}{Alex Kendall} {and} \bibinfo{person}{Yarin Gal}.} \bibinfo{year}{2017}\natexlab{}.
\newblock \showarticletitle{What uncertainties do we need in bayesian deep learning for computer vision?}
\newblock \bibinfo{journal}{\emph{Advances in neural information processing systems}}  \bibinfo{volume}{30} (\bibinfo{year}{2017}).
\newblock


\bibitem[\protect\citeauthoryear{Kipf and Welling}{Kipf and Welling}{2016}]%
        {kipf2016semi}
\bibfield{author}{\bibinfo{person}{Thomas~N Kipf} {and} \bibinfo{person}{Max Welling}.} \bibinfo{year}{2016}\natexlab{}.
\newblock \showarticletitle{Semi-supervised classification with graph convolutional networks}.
\newblock \bibinfo{journal}{\emph{arXiv preprint arXiv:1609.02907}} (\bibinfo{year}{2016}).
\newblock


\bibitem[\protect\citeauthoryear{Krügel, Toth, and Kirda}{Krügel et~al\mbox{.}}{2002}]%
        {Christopher2002Service}
\bibfield{author}{\bibinfo{person}{Christopher Krügel}, \bibinfo{person}{Thomas Toth}, {and} \bibinfo{person}{Engin Kirda}.} \bibinfo{year}{2002}\natexlab{}.
\newblock \showarticletitle{Service specific anomaly detection for network intrusion detection}. In \bibinfo{booktitle}{\emph{the 2002 ACM symposium}}.
\newblock


\bibitem[\protect\citeauthoryear{Leskovec, Kleinberg, and Faloutsos}{Leskovec et~al\mbox{.}}{2005}]%
        {leskovec2005graphs}
\bibfield{author}{\bibinfo{person}{Jure Leskovec}, \bibinfo{person}{Jon Kleinberg}, {and} \bibinfo{person}{Christos Faloutsos}.} \bibinfo{year}{2005}\natexlab{}.
\newblock \showarticletitle{Graphs over time: densification laws, shrinking diameters and possible explanations}. In \bibinfo{booktitle}{\emph{Proceedings of the eleventh ACM SIGKDD international conference on Knowledge discovery in data mining}}. \bibinfo{pages}{177--187}.
\newblock


\bibitem[\protect\citeauthoryear{Lim, Hohne, Li, Huang, Gupta, Bhalerao, and Lim}{Lim et~al\mbox{.}}{2021}]%
        {lim2021large}
\bibfield{author}{\bibinfo{person}{Derek Lim}, \bibinfo{person}{Felix Hohne}, \bibinfo{person}{Xiuyu Li}, \bibinfo{person}{Sijia~Linda Huang}, \bibinfo{person}{Vaishnavi Gupta}, \bibinfo{person}{Omkar Bhalerao}, {and} \bibinfo{person}{Ser~Nam Lim}.} \bibinfo{year}{2021}\natexlab{}.
\newblock \showarticletitle{Large scale learning on non-homophilous graphs: New benchmarks and strong simple methods}.
\newblock \bibinfo{journal}{\emph{Advances in Neural Information Processing Systems}}  \bibinfo{volume}{34} (\bibinfo{year}{2021}), \bibinfo{pages}{20887--20902}.
\newblock


\bibitem[\protect\citeauthoryear{Lin, Chen, Wang, Xi, Qu, Dai, Zhang, Tang, Yu, and Zhang}{Lin et~al\mbox{.}}{2024}]%
        {lin2024clickprompt}
\bibfield{author}{\bibinfo{person}{Jianghao Lin}, \bibinfo{person}{Bo Chen}, \bibinfo{person}{Hangyu Wang}, \bibinfo{person}{Yunjia Xi}, \bibinfo{person}{Yanru Qu}, \bibinfo{person}{Xinyi Dai}, \bibinfo{person}{Kangning Zhang}, \bibinfo{person}{Ruiming Tang}, \bibinfo{person}{Yong Yu}, {and} \bibinfo{person}{Weinan Zhang}.} \bibinfo{year}{2024}\natexlab{}.
\newblock \showarticletitle{ClickPrompt: CTR Models are Strong Prompt Generators for Adapting Language Models to CTR Prediction}. In \bibinfo{booktitle}{\emph{Proceedings of the ACM on Web Conference 2024}}. \bibinfo{pages}{3319--3330}.
\newblock


\bibitem[\protect\citeauthoryear{Liu, Li, Wu, and Lee}{Liu et~al\mbox{.}}{2023}]%
        {liu2023visual}
\bibfield{author}{\bibinfo{person}{Haotian Liu}, \bibinfo{person}{Chunyuan Li}, \bibinfo{person}{Qingyang Wu}, {and} \bibinfo{person}{Yong~Jae Lee}.} \bibinfo{year}{2023}\natexlab{}.
\newblock \showarticletitle{Visual instruction tuning}.
\newblock \bibinfo{journal}{\emph{arXiv preprint arXiv:2304.08485}} (\bibinfo{year}{2023}).
\newblock


\bibitem[\protect\citeauthoryear{Liu, Ao, Feng, and He}{Liu et~al\mbox{.}}{2022}]%
        {liu2022ud}
\bibfield{author}{\bibinfo{person}{Yang Liu}, \bibinfo{person}{Xiang Ao}, \bibinfo{person}{Fuli Feng}, {and} \bibinfo{person}{Qing He}.} \bibinfo{year}{2022}\natexlab{}.
\newblock \showarticletitle{UD-GNN: Uncertainty-aware Debiased Training on Semi-Homophilous Graphs}. In \bibinfo{booktitle}{\emph{Proceedings of the 28th ACM SIGKDD Conference on Knowledge Discovery and Data Mining}}. \bibinfo{pages}{1131--1140}.
\newblock


\bibitem[\protect\citeauthoryear{Liu, Ao, Qin, Chi, Feng, Yang, and He}{Liu et~al\mbox{.}}{2021}]%
        {liu2021pick}
\bibfield{author}{\bibinfo{person}{Yang Liu}, \bibinfo{person}{Xiang Ao}, \bibinfo{person}{Zidi Qin}, \bibinfo{person}{Jianfeng Chi}, \bibinfo{person}{Jinghua Feng}, \bibinfo{person}{Hao Yang}, {and} \bibinfo{person}{Qing He}.} \bibinfo{year}{2021}\natexlab{}.
\newblock \showarticletitle{Pick and choose: a GNN-based imbalanced learning approach for fraud detection}. In \bibinfo{booktitle}{\emph{Proceedings of the web conference 2021}}. \bibinfo{pages}{3168--3177}.
\newblock


\bibitem[\protect\citeauthoryear{Liu, Zhang, Li, Yan, Gao, Chen, Yuan, Huang, Sun, Gao, et~al\mbox{.}}{Liu et~al\mbox{.}}{2024}]%
        {liu2024sora}
\bibfield{author}{\bibinfo{person}{Yixin Liu}, \bibinfo{person}{Kai Zhang}, \bibinfo{person}{Yuan Li}, \bibinfo{person}{Zhiling Yan}, \bibinfo{person}{Chujie Gao}, \bibinfo{person}{Ruoxi Chen}, \bibinfo{person}{Zhengqing Yuan}, \bibinfo{person}{Yue Huang}, \bibinfo{person}{Hanchi Sun}, \bibinfo{person}{Jianfeng Gao}, {et~al\mbox{.}}} \bibinfo{year}{2024}\natexlab{}.
\newblock \showarticletitle{Sora: A Review on Background, Technology, Limitations, and Opportunities of Large Vision Models}.
\newblock \bibinfo{journal}{\emph{arXiv preprint arXiv:2402.17177}} (\bibinfo{year}{2024}).
\newblock


\bibitem[\protect\citeauthoryear{Lo, Rensi, Torng, and Altman}{Lo et~al\mbox{.}}{2018}]%
        {lo2018machine}
\bibfield{author}{\bibinfo{person}{Yu-Chen Lo}, \bibinfo{person}{Stefano~E Rensi}, \bibinfo{person}{Wen Torng}, {and} \bibinfo{person}{Russ~B Altman}.} \bibinfo{year}{2018}\natexlab{}.
\newblock \showarticletitle{Machine learning in chemoinformatics and drug discovery}.
\newblock \bibinfo{journal}{\emph{Drug discovery today}} \bibinfo{volume}{23}, \bibinfo{number}{8} (\bibinfo{year}{2018}), \bibinfo{pages}{1538--1546}.
\newblock


\bibitem[\protect\citeauthoryear{Luan, Zhao, Hua, Chang, and Precup}{Luan et~al\mbox{.}}{2020}]%
        {luan2020complete}
\bibfield{author}{\bibinfo{person}{Sitao Luan}, \bibinfo{person}{Mingde Zhao}, \bibinfo{person}{Chenqing Hua}, \bibinfo{person}{Xiao-Wen Chang}, {and} \bibinfo{person}{Doina Precup}.} \bibinfo{year}{2020}\natexlab{}.
\newblock \showarticletitle{Complete the missing half: Augmenting aggregation filtering with diversification for graph convolutional networks}.
\newblock \bibinfo{journal}{\emph{arXiv preprint arXiv:2008.08844}} (\bibinfo{year}{2020}).
\newblock


\bibitem[\protect\citeauthoryear{Ma, Zhou, Cui, Yang, and Zhu}{Ma et~al\mbox{.}}{2019}]%
        {ma2019learning}
\bibfield{author}{\bibinfo{person}{Jianxin Ma}, \bibinfo{person}{Chang Zhou}, \bibinfo{person}{Peng Cui}, \bibinfo{person}{Hongxia Yang}, {and} \bibinfo{person}{Wenwu Zhu}.} \bibinfo{year}{2019}\natexlab{}.
\newblock \showarticletitle{Learning disentangled representations for recommendation}.
\newblock \bibinfo{journal}{\emph{Advances in neural information processing systems}}  \bibinfo{volume}{32} (\bibinfo{year}{2019}).
\newblock


\bibitem[\protect\citeauthoryear{Mccallum, Wang, and Corrada-Emmanuel}{Mccallum et~al\mbox{.}}{2007}]%
        {2007Topic}
\bibfield{author}{\bibinfo{person}{Andrew Mccallum}, \bibinfo{person}{Xuerui Wang}, {and} \bibinfo{person}{Andrés Corrada-Emmanuel}.} \bibinfo{year}{2007}\natexlab{}.
\newblock \showarticletitle{Topic and Role Discovery in Social Networks with Experiments on Enron and Academic Email}.
\newblock \bibinfo{journal}{\emph{Journal of Artificial Intelligence Research}} \bibinfo{volume}{30}, \bibinfo{number}{1} (\bibinfo{year}{2007}), \bibinfo{pages}{249--272}.
\newblock


\bibitem[\protect\citeauthoryear{McCallum, Nigam, Rennie, and Seymore}{McCallum et~al\mbox{.}}{2000}]%
        {mccallum2000automating}
\bibfield{author}{\bibinfo{person}{Andrew~Kachites McCallum}, \bibinfo{person}{Kamal Nigam}, \bibinfo{person}{Jason Rennie}, {and} \bibinfo{person}{Kristie Seymore}.} \bibinfo{year}{2000}\natexlab{}.
\newblock \showarticletitle{Automating the construction of internet portals with machine learning}.
\newblock \bibinfo{journal}{\emph{Information Retrieval}}  \bibinfo{volume}{3} (\bibinfo{year}{2000}), \bibinfo{pages}{127--163}.
\newblock


\bibitem[\protect\citeauthoryear{Mysore, McCallum, and Zamani}{Mysore et~al\mbox{.}}{2023}]%
        {mysore2023large}
\bibfield{author}{\bibinfo{person}{Sheshera Mysore}, \bibinfo{person}{Andrew McCallum}, {and} \bibinfo{person}{Hamed Zamani}.} \bibinfo{year}{2023}\natexlab{}.
\newblock \showarticletitle{Large Language Model Augmented Narrative Driven Recommendations}.
\newblock \bibinfo{journal}{\emph{arXiv preprint arXiv:2306.02250}} (\bibinfo{year}{2023}).
\newblock


\bibitem[\protect\citeauthoryear{Pan, Luo, Wang, Chen, Wang, and Wu}{Pan et~al\mbox{.}}{2024}]%
        {pan2024unifying}
\bibfield{author}{\bibinfo{person}{Shirui Pan}, \bibinfo{person}{Linhao Luo}, \bibinfo{person}{Yufei Wang}, \bibinfo{person}{Chen Chen}, \bibinfo{person}{Jiapu Wang}, {and} \bibinfo{person}{Xindong Wu}.} \bibinfo{year}{2024}\natexlab{}.
\newblock \showarticletitle{Unifying large language models and knowledge graphs: A roadmap}.
\newblock \bibinfo{journal}{\emph{IEEE Transactions on Knowledge and Data Engineering}} (\bibinfo{year}{2024}).
\newblock


\bibitem[\protect\citeauthoryear{Park, Song, and Yang}{Park et~al\mbox{.}}{2021}]%
        {park2021graphens}
\bibfield{author}{\bibinfo{person}{Joonhyung Park}, \bibinfo{person}{Jaeyun Song}, {and} \bibinfo{person}{Eunho Yang}.} \bibinfo{year}{2021}\natexlab{}.
\newblock \showarticletitle{Graphens: Neighbor-aware ego network synthesis for class-imbalanced node classification}. In \bibinfo{booktitle}{\emph{International conference on learning representations}}.
\newblock


\bibitem[\protect\citeauthoryear{Pei, Wei, Chang, Lei, and Yang}{Pei et~al\mbox{.}}{2019}]%
        {pei2019geom}
\bibfield{author}{\bibinfo{person}{Hongbin Pei}, \bibinfo{person}{Bingzhe Wei}, \bibinfo{person}{Kevin Chen-Chuan Chang}, \bibinfo{person}{Yu Lei}, {and} \bibinfo{person}{Bo Yang}.} \bibinfo{year}{2019}\natexlab{}.
\newblock \showarticletitle{Geom-GCN: Geometric Graph Convolutional Networks}. In \bibinfo{booktitle}{\emph{International Conference on Learning Representations}}.
\newblock


\bibitem[\protect\citeauthoryear{Pourhabibi, Ong, Kam, and Boo}{Pourhabibi et~al\mbox{.}}{2020}]%
        {pourhabibi2020fraud}
\bibfield{author}{\bibinfo{person}{Tahereh Pourhabibi}, \bibinfo{person}{Kok-Leong Ong}, \bibinfo{person}{Booi~H Kam}, {and} \bibinfo{person}{Yee~Ling Boo}.} \bibinfo{year}{2020}\natexlab{}.
\newblock \showarticletitle{Fraud detection: A systematic literature review of graph-based anomaly detection approaches}.
\newblock \bibinfo{journal}{\emph{Decision Support Systems}}  \bibinfo{volume}{133} (\bibinfo{year}{2020}), \bibinfo{pages}{113303}.
\newblock


\bibitem[\protect\citeauthoryear{Qu, Zhu, Zheng, Shi, and Yin}{Qu et~al\mbox{.}}{2021}]%
        {qu2021imgagn}
\bibfield{author}{\bibinfo{person}{Liang Qu}, \bibinfo{person}{Huaisheng Zhu}, \bibinfo{person}{Ruiqi Zheng}, \bibinfo{person}{Yuhui Shi}, {and} \bibinfo{person}{Hongzhi Yin}.} \bibinfo{year}{2021}\natexlab{}.
\newblock \showarticletitle{Imgagn: Imbalanced network embedding via generative adversarial graph networks}. In \bibinfo{booktitle}{\emph{Proceedings of the 27th ACM SIGKDD Conference on Knowledge Discovery \& Data Mining}}. \bibinfo{pages}{1390--1398}.
\newblock


\bibitem[\protect\citeauthoryear{Radford, Wu, Child, Luan, Amodei, Sutskever, et~al\mbox{.}}{Radford et~al\mbox{.}}{2019}]%
        {radford2019language}
\bibfield{author}{\bibinfo{person}{Alec Radford}, \bibinfo{person}{Jeffrey Wu}, \bibinfo{person}{Rewon Child}, \bibinfo{person}{David Luan}, \bibinfo{person}{Dario Amodei}, \bibinfo{person}{Ilya Sutskever}, {et~al\mbox{.}}} \bibinfo{year}{2019}\natexlab{}.
\newblock \showarticletitle{Language models are unsupervised multitask learners}.
\newblock \bibinfo{journal}{\emph{OpenAI blog}} \bibinfo{volume}{1}, \bibinfo{number}{8} (\bibinfo{year}{2019}), \bibinfo{pages}{9}.
\newblock


\bibitem[\protect\citeauthoryear{Reid, Savinov, Teplyashin, Lepikhin, Lillicrap, Alayrac, Soricut, Lazaridou, Firat, Schrittwieser, et~al\mbox{.}}{Reid et~al\mbox{.}}{2024}]%
        {reid2024gemini}
\bibfield{author}{\bibinfo{person}{Machel Reid}, \bibinfo{person}{Nikolay Savinov}, \bibinfo{person}{Denis Teplyashin}, \bibinfo{person}{Dmitry Lepikhin}, \bibinfo{person}{Timothy Lillicrap}, \bibinfo{person}{Jean-baptiste Alayrac}, \bibinfo{person}{Radu Soricut}, \bibinfo{person}{Angeliki Lazaridou}, \bibinfo{person}{Orhan Firat}, \bibinfo{person}{Julian Schrittwieser}, {et~al\mbox{.}}} \bibinfo{year}{2024}\natexlab{}.
\newblock \showarticletitle{Gemini 1.5: Unlocking multimodal understanding across millions of tokens of context}.
\newblock \bibinfo{journal}{\emph{arXiv preprint arXiv:2403.05530}} (\bibinfo{year}{2024}).
\newblock


\bibitem[\protect\citeauthoryear{Ren, Wei, Xia, Su, Cheng, Wang, Yin, and Huang}{Ren et~al\mbox{.}}{2023}]%
        {ren2023representation}
\bibfield{author}{\bibinfo{person}{Xubin Ren}, \bibinfo{person}{Wei Wei}, \bibinfo{person}{Lianghao Xia}, \bibinfo{person}{Lixin Su}, \bibinfo{person}{Suqi Cheng}, \bibinfo{person}{Junfeng Wang}, \bibinfo{person}{Dawei Yin}, {and} \bibinfo{person}{Chao Huang}.} \bibinfo{year}{2023}\natexlab{}.
\newblock \showarticletitle{Representation learning with large language models for recommendation}.
\newblock \bibinfo{journal}{\emph{arXiv preprint arXiv:2310.15950}} (\bibinfo{year}{2023}).
\newblock


\bibitem[\protect\citeauthoryear{Riesen and Bunke}{Riesen and Bunke}{2008}]%
        {riesen2008iam}
\bibfield{author}{\bibinfo{person}{Kaspar Riesen} {and} \bibinfo{person}{Horst Bunke}.} \bibinfo{year}{2008}\natexlab{}.
\newblock \showarticletitle{IAM graph database repository for graph based pattern recognition and machine learning}. In \bibinfo{booktitle}{\emph{Structural, Syntactic, and Statistical Pattern Recognition: Joint IAPR International Workshop, SSPR \& SPR 2008, Orlando, USA, December 4-6, 2008. Proceedings}}. Springer, \bibinfo{pages}{287--297}.
\newblock


\bibitem[\protect\citeauthoryear{Rogers and Bhowmik}{Rogers and Bhowmik}{1970a}]%
        {1970Homophily}
\bibfield{author}{\bibinfo{person}{E.~M. Rogers} {and} \bibinfo{person}{D.~K. Bhowmik}.} \bibinfo{year}{1970}\natexlab{a}.
\newblock \showarticletitle{Homophily-heterophily: Relational concepts for communication research}.
\newblock \bibinfo{journal}{\emph{Public Opinion Quarterly}} \bibinfo{volume}{34}, \bibinfo{number}{4} (\bibinfo{year}{1970}), \bibinfo{pages}{523--538}.
\newblock


\bibitem[\protect\citeauthoryear{Rogers and Bhowmik}{Rogers and Bhowmik}{1970b}]%
        {rogers1970homophily}
\bibfield{author}{\bibinfo{person}{Everett~M Rogers} {and} \bibinfo{person}{Dilip~K Bhowmik}.} \bibinfo{year}{1970}\natexlab{b}.
\newblock \showarticletitle{Homophily-heterophily: Relational concepts for communication research}.
\newblock \bibinfo{journal}{\emph{Public opinion quarterly}} \bibinfo{volume}{34}, \bibinfo{number}{4} (\bibinfo{year}{1970}), \bibinfo{pages}{523--538}.
\newblock


\bibitem[\protect\citeauthoryear{Sen, Namata, Bilgic, Getoor, Galligher, and Eliassi-Rad}{Sen et~al\mbox{.}}{2008}]%
        {sen2008collective}
\bibfield{author}{\bibinfo{person}{Prithviraj Sen}, \bibinfo{person}{Galileo Namata}, \bibinfo{person}{Mustafa Bilgic}, \bibinfo{person}{Lise Getoor}, \bibinfo{person}{Brian Galligher}, {and} \bibinfo{person}{Tina Eliassi-Rad}.} \bibinfo{year}{2008}\natexlab{}.
\newblock \showarticletitle{Collective classification in network data}.
\newblock \bibinfo{journal}{\emph{AI magazine}} \bibinfo{volume}{29}, \bibinfo{number}{3} (\bibinfo{year}{2008}), \bibinfo{pages}{93--93}.
\newblock


\bibitem[\protect\citeauthoryear{Shi, Hu, Zhao, and Philip}{Shi et~al\mbox{.}}{2018}]%
        {shi2018heterogeneous}
\bibfield{author}{\bibinfo{person}{Chuan Shi}, \bibinfo{person}{Binbin Hu}, \bibinfo{person}{Wayne~Xin Zhao}, {and} \bibinfo{person}{S~Yu Philip}.} \bibinfo{year}{2018}\natexlab{}.
\newblock \showarticletitle{Heterogeneous information network embedding for recommendation}.
\newblock \bibinfo{journal}{\emph{IEEE transactions on knowledge and data engineering}} \bibinfo{volume}{31}, \bibinfo{number}{2} (\bibinfo{year}{2018}), \bibinfo{pages}{357--370}.
\newblock


\bibitem[\protect\citeauthoryear{Song, Park, and Yang}{Song et~al\mbox{.}}{2022}]%
        {song2022tam}
\bibfield{author}{\bibinfo{person}{Jaeyun Song}, \bibinfo{person}{Joonhyung Park}, {and} \bibinfo{person}{Eunho Yang}.} \bibinfo{year}{2022}\natexlab{}.
\newblock \showarticletitle{TAM: topology-aware margin loss for class-imbalanced node classification}. In \bibinfo{booktitle}{\emph{International Conference on Machine Learning}}. PMLR, \bibinfo{pages}{20369--20383}.
\newblock


\bibitem[\protect\citeauthoryear{St{\"a}rk, Beaini, Corso, Tossou, Dallago, G{\"u}nnemann, and Li{\`o}}{St{\"a}rk et~al\mbox{.}}{2022}]%
        {stark20223d}
\bibfield{author}{\bibinfo{person}{Hannes St{\"a}rk}, \bibinfo{person}{Dominique Beaini}, \bibinfo{person}{Gabriele Corso}, \bibinfo{person}{Prudencio Tossou}, \bibinfo{person}{Christian Dallago}, \bibinfo{person}{Stephan G{\"u}nnemann}, {and} \bibinfo{person}{Pietro Li{\`o}}.} \bibinfo{year}{2022}\natexlab{}.
\newblock \showarticletitle{3d infomax improves gnns for molecular property prediction}. In \bibinfo{booktitle}{\emph{International Conference on Machine Learning}}. PMLR, \bibinfo{pages}{20479--20502}.
\newblock


\bibitem[\protect\citeauthoryear{Sun, Ren, Ma, and Zhang}{Sun et~al\mbox{.}}{2023}]%
        {sun2023large}
\bibfield{author}{\bibinfo{person}{Shengyin Sun}, \bibinfo{person}{Yuxiang Ren}, \bibinfo{person}{Chen Ma}, {and} \bibinfo{person}{Xuecang Zhang}.} \bibinfo{year}{2023}\natexlab{}.
\newblock \showarticletitle{Large Language Models as Topological Structure Enhancers for Text-Attributed Graphs}.
\newblock \bibinfo{journal}{\emph{arXiv preprint arXiv:2311.14324}} (\bibinfo{year}{2023}).
\newblock


\bibitem[\protect\citeauthoryear{Tang, Yang, Wei, Shi, Su, Cheng, Yin, and Huang}{Tang et~al\mbox{.}}{2023}]%
        {tang2023graphgpt}
\bibfield{author}{\bibinfo{person}{Jiabin Tang}, \bibinfo{person}{Yuhao Yang}, \bibinfo{person}{Wei Wei}, \bibinfo{person}{Lei Shi}, \bibinfo{person}{Lixin Su}, \bibinfo{person}{Suqi Cheng}, \bibinfo{person}{Dawei Yin}, {and} \bibinfo{person}{Chao Huang}.} \bibinfo{year}{2023}\natexlab{}.
\newblock \showarticletitle{Graphgpt: Graph instruction tuning for large language models}.
\newblock \bibinfo{journal}{\emph{arXiv preprint arXiv:2310.13023}} (\bibinfo{year}{2023}).
\newblock


\bibitem[\protect\citeauthoryear{Tian, Liu, Wang, and Zhou}{Tian et~al\mbox{.}}{2023}]%
        {tian2023asa}
\bibfield{author}{\bibinfo{person}{Yue Tian}, \bibinfo{person}{Guanjun Liu}, \bibinfo{person}{Jiacun Wang}, {and} \bibinfo{person}{Mengchu Zhou}.} \bibinfo{year}{2023}\natexlab{}.
\newblock \showarticletitle{ASA-GNN: Adaptive Sampling and Aggregation-Based Graph Neural Network for Transaction Fraud Detection}.
\newblock \bibinfo{journal}{\emph{IEEE Transactions on Computational Social Systems}} (\bibinfo{year}{2023}).
\newblock


\bibitem[\protect\citeauthoryear{Touvron, Martin, Stone, Albert, Almahairi, Babaei, Bashlykov, Batra, Bhargava, Bhosale, et~al\mbox{.}}{Touvron et~al\mbox{.}}{2023}]%
        {touvron2023llama}
\bibfield{author}{\bibinfo{person}{Hugo Touvron}, \bibinfo{person}{Louis Martin}, \bibinfo{person}{Kevin Stone}, \bibinfo{person}{Peter Albert}, \bibinfo{person}{Amjad Almahairi}, \bibinfo{person}{Yasmine Babaei}, \bibinfo{person}{Nikolay Bashlykov}, \bibinfo{person}{Soumya Batra}, \bibinfo{person}{Prajjwal Bhargava}, \bibinfo{person}{Shruti Bhosale}, {et~al\mbox{.}}} \bibinfo{year}{2023}\natexlab{}.
\newblock \showarticletitle{Llama 2: Open foundation and fine-tuned chat models}.
\newblock \bibinfo{journal}{\emph{arXiv preprint arXiv:2307.09288}} (\bibinfo{year}{2023}).
\newblock


\bibitem[\protect\citeauthoryear{Velickovic, Cucurull, Casanova, Romero, Lio, Bengio, et~al\mbox{.}}{Velickovic et~al\mbox{.}}{2017}]%
        {velickovic2017graph}
\bibfield{author}{\bibinfo{person}{Petar Velickovic}, \bibinfo{person}{Guillem Cucurull}, \bibinfo{person}{Arantxa Casanova}, \bibinfo{person}{Adriana Romero}, \bibinfo{person}{Pietro Lio}, \bibinfo{person}{Yoshua Bengio}, {et~al\mbox{.}}} \bibinfo{year}{2017}\natexlab{}.
\newblock \showarticletitle{Graph attention networks}.
\newblock \bibinfo{journal}{\emph{stat}} \bibinfo{volume}{1050}, \bibinfo{number}{20} (\bibinfo{year}{2017}), \bibinfo{pages}{10--48550}.
\newblock


\bibitem[\protect\citeauthoryear{Wang, Feng, He, Tan, Han, and Tsvetkov}{Wang et~al\mbox{.}}{2023a}]%
        {wang2023can}
\bibfield{author}{\bibinfo{person}{Heng Wang}, \bibinfo{person}{Shangbin Feng}, \bibinfo{person}{Tianxing He}, \bibinfo{person}{Zhaoxuan Tan}, \bibinfo{person}{Xiaochuang Han}, {and} \bibinfo{person}{Yulia Tsvetkov}.} \bibinfo{year}{2023}\natexlab{a}.
\newblock \showarticletitle{Can Language Models Solve Graph Problems in Natural Language?}
\newblock \bibinfo{journal}{\emph{arXiv preprint arXiv:2305.10037}} (\bibinfo{year}{2023}).
\newblock


\bibitem[\protect\citeauthoryear{Wang, Fu, Du, Gao, Huang, Liu, Chandak, Liu, Van~Katwyk, Deac, et~al\mbox{.}}{Wang et~al\mbox{.}}{2023b}]%
        {wang2023scientific}
\bibfield{author}{\bibinfo{person}{Hanchen Wang}, \bibinfo{person}{Tianfan Fu}, \bibinfo{person}{Yuanqi Du}, \bibinfo{person}{Wenhao Gao}, \bibinfo{person}{Kexin Huang}, \bibinfo{person}{Ziming Liu}, \bibinfo{person}{Payal Chandak}, \bibinfo{person}{Shengchao Liu}, \bibinfo{person}{Peter Van~Katwyk}, \bibinfo{person}{Andreea Deac}, {et~al\mbox{.}}} \bibinfo{year}{2023}\natexlab{b}.
\newblock \showarticletitle{Scientific discovery in the age of artificial intelligence}.
\newblock \bibinfo{journal}{\emph{Nature}} \bibinfo{volume}{620}, \bibinfo{number}{7972} (\bibinfo{year}{2023}), \bibinfo{pages}{47--60}.
\newblock


\bibitem[\protect\citeauthoryear{Wang, Bo, Shi, Fan, Ye, and Philip}{Wang et~al\mbox{.}}{2022}]%
        {wang2022survey}
\bibfield{author}{\bibinfo{person}{Xiao Wang}, \bibinfo{person}{Deyu Bo}, \bibinfo{person}{Chuan Shi}, \bibinfo{person}{Shaohua Fan}, \bibinfo{person}{Yanfang Ye}, {and} \bibinfo{person}{S~Yu Philip}.} \bibinfo{year}{2022}\natexlab{}.
\newblock \showarticletitle{A survey on heterogeneous graph embedding: methods, techniques, applications and sources}.
\newblock \bibinfo{journal}{\emph{IEEE Transactions on Big Data}} \bibinfo{volume}{9}, \bibinfo{number}{2} (\bibinfo{year}{2022}), \bibinfo{pages}{415--436}.
\newblock


\bibitem[\protect\citeauthoryear{Wang, Ji, Shi, Wang, Ye, Cui, and Yu}{Wang et~al\mbox{.}}{2019}]%
        {wang2019heterogeneous}
\bibfield{author}{\bibinfo{person}{Xiao Wang}, \bibinfo{person}{Houye Ji}, \bibinfo{person}{Chuan Shi}, \bibinfo{person}{Bai Wang}, \bibinfo{person}{Yanfang Ye}, \bibinfo{person}{Peng Cui}, {and} \bibinfo{person}{Philip~S Yu}.} \bibinfo{year}{2019}\natexlab{}.
\newblock \showarticletitle{Heterogeneous graph attention network}. In \bibinfo{booktitle}{\emph{The world wide web conference}}. \bibinfo{pages}{2022--2032}.
\newblock


\bibitem[\protect\citeauthoryear{Wei, Tay, Bommasani, Raffel, Zoph, Borgeaud, Yogatama, Bosma, Zhou, Metzler, et~al\mbox{.}}{Wei et~al\mbox{.}}{2022}]%
        {wei2022emergent}
\bibfield{author}{\bibinfo{person}{Jason Wei}, \bibinfo{person}{Yi Tay}, \bibinfo{person}{Rishi Bommasani}, \bibinfo{person}{Colin Raffel}, \bibinfo{person}{Barret Zoph}, \bibinfo{person}{Sebastian Borgeaud}, \bibinfo{person}{Dani Yogatama}, \bibinfo{person}{Maarten Bosma}, \bibinfo{person}{Denny Zhou}, \bibinfo{person}{Donald Metzler}, {et~al\mbox{.}}} \bibinfo{year}{2022}\natexlab{}.
\newblock \showarticletitle{Emergent abilities of large language models}.
\newblock \bibinfo{journal}{\emph{arXiv preprint arXiv:2206.07682}} (\bibinfo{year}{2022}).
\newblock


\bibitem[\protect\citeauthoryear{Wu, Souza, Zhang, Fifty, Yu, and Weinberger}{Wu et~al\mbox{.}}{2019}]%
        {wu2019simplifying}
\bibfield{author}{\bibinfo{person}{Felix Wu}, \bibinfo{person}{Amauri Souza}, \bibinfo{person}{Tianyi Zhang}, \bibinfo{person}{Christopher Fifty}, \bibinfo{person}{Tao Yu}, {and} \bibinfo{person}{Kilian Weinberger}.} \bibinfo{year}{2019}\natexlab{}.
\newblock \showarticletitle{Simplifying graph convolutional networks}. In \bibinfo{booktitle}{\emph{International conference on machine learning}}. PMLR, \bibinfo{pages}{6861--6871}.
\newblock


\bibitem[\protect\citeauthoryear{Xu, Li, Tian, Sonobe, Kawarabayashi, and Jegelka}{Xu et~al\mbox{.}}{2018}]%
        {xu2018representation}
\bibfield{author}{\bibinfo{person}{Keyulu Xu}, \bibinfo{person}{Chengtao Li}, \bibinfo{person}{Yonglong Tian}, \bibinfo{person}{Tomohiro Sonobe}, \bibinfo{person}{Ken-ichi Kawarabayashi}, {and} \bibinfo{person}{Stefanie Jegelka}.} \bibinfo{year}{2018}\natexlab{}.
\newblock \showarticletitle{Representation learning on graphs with jumping knowledge networks}. In \bibinfo{booktitle}{\emph{International conference on machine learning}}. PMLR, \bibinfo{pages}{5453--5462}.
\newblock


\bibitem[\protect\citeauthoryear{Ye, Zhang, Wang, Xu, and Zhang}{Ye et~al\mbox{.}}{2023}]%
        {ye2023natural}
\bibfield{author}{\bibinfo{person}{Ruosong Ye}, \bibinfo{person}{Caiqi Zhang}, \bibinfo{person}{Runhui Wang}, \bibinfo{person}{Shuyuan Xu}, {and} \bibinfo{person}{Yongfeng Zhang}.} \bibinfo{year}{2023}\natexlab{}.
\newblock \showarticletitle{Natural language is all a graph needs}.
\newblock \bibinfo{journal}{\emph{arXiv preprint arXiv:2308.07134}} (\bibinfo{year}{2023}).
\newblock


\bibitem[\protect\citeauthoryear{Ying, He, Chen, Eksombatchai, Hamilton, and Leskovec}{Ying et~al\mbox{.}}{2018}]%
        {ying2018graph}
\bibfield{author}{\bibinfo{person}{Rex Ying}, \bibinfo{person}{Ruining He}, \bibinfo{person}{Kaifeng Chen}, \bibinfo{person}{Pong Eksombatchai}, \bibinfo{person}{William~L Hamilton}, {and} \bibinfo{person}{Jure Leskovec}.} \bibinfo{year}{2018}\natexlab{}.
\newblock \showarticletitle{Graph convolutional neural networks for web-scale recommender systems}. In \bibinfo{booktitle}{\emph{Proceedings of the 24th ACM SIGKDD international conference on knowledge discovery \& data mining}}. \bibinfo{pages}{974--983}.
\newblock


\bibitem[\protect\citeauthoryear{Yun, Jeong, Kim, Kang, and Kim}{Yun et~al\mbox{.}}{2019}]%
        {yun2019graph}
\bibfield{author}{\bibinfo{person}{Seongjun Yun}, \bibinfo{person}{Minbyul Jeong}, \bibinfo{person}{Raehyun Kim}, \bibinfo{person}{Jaewoo Kang}, {and} \bibinfo{person}{Hyunwoo~J Kim}.} \bibinfo{year}{2019}\natexlab{}.
\newblock \showarticletitle{Graph transformer networks}.
\newblock \bibinfo{journal}{\emph{Advances in neural information processing systems}}  \bibinfo{volume}{32} (\bibinfo{year}{2019}).
\newblock


\bibitem[\protect\citeauthoryear{Zhang, Song, Huang, Swami, and Chawla}{Zhang et~al\mbox{.}}{2019}]%
        {zhang2019heterogeneous}
\bibfield{author}{\bibinfo{person}{Chuxu Zhang}, \bibinfo{person}{Dongjin Song}, \bibinfo{person}{Chao Huang}, \bibinfo{person}{Ananthram Swami}, {and} \bibinfo{person}{Nitesh~V Chawla}.} \bibinfo{year}{2019}\natexlab{}.
\newblock \showarticletitle{Heterogeneous graph neural network}. In \bibinfo{booktitle}{\emph{Proceedings of the 25th ACM SIGKDD international conference on knowledge discovery \& data mining}}. \bibinfo{pages}{793--803}.
\newblock


\bibitem[\protect\citeauthoryear{Zhang and Zitnik}{Zhang and Zitnik}{2020}]%
        {zhang2020gnnguard}
\bibfield{author}{\bibinfo{person}{Xiang Zhang} {and} \bibinfo{person}{Marinka Zitnik}.} \bibinfo{year}{2020}\natexlab{}.
\newblock \showarticletitle{Gnnguard: Defending graph neural networks against adversarial attacks}.
\newblock \bibinfo{journal}{\emph{Advances in neural information processing systems}}  \bibinfo{volume}{33} (\bibinfo{year}{2020}), \bibinfo{pages}{9263--9275}.
\newblock


\bibitem[\protect\citeauthoryear{Zhang, Wang, Zhang, Shen, Shen, and Zhu}{Zhang et~al\mbox{.}}{2023}]%
        {zhang2023unsupervised}
\bibfield{author}{\bibinfo{person}{Zeyang Zhang}, \bibinfo{person}{Xin Wang}, \bibinfo{person}{Ziwei Zhang}, \bibinfo{person}{Guangyao Shen}, \bibinfo{person}{Shiqi Shen}, {and} \bibinfo{person}{Wenwu Zhu}.} \bibinfo{year}{2023}\natexlab{}.
\newblock \showarticletitle{Unsupervised graph neural architecture search with disentangled self-supervision}. In \bibinfo{booktitle}{\emph{Thirty-seventh Conference on Neural Information Processing Systems}}.
\newblock


\bibitem[\protect\citeauthoryear{Zhao, Zhuo, Shen, Qu, Liu, Bronstein, Zhu, and Tang}{Zhao et~al\mbox{.}}{2023b}]%
        {zhao2023graphtext}
\bibfield{author}{\bibinfo{person}{Jianan Zhao}, \bibinfo{person}{Le Zhuo}, \bibinfo{person}{Yikang Shen}, \bibinfo{person}{Meng Qu}, \bibinfo{person}{Kai Liu}, \bibinfo{person}{Michael Bronstein}, \bibinfo{person}{Zhaocheng Zhu}, {and} \bibinfo{person}{Jian Tang}.} \bibinfo{year}{2023}\natexlab{b}.
\newblock \showarticletitle{Graphtext: Graph reasoning in text space}.
\newblock \bibinfo{journal}{\emph{arXiv preprint arXiv:2310.01089}} (\bibinfo{year}{2023}).
\newblock


\bibitem[\protect\citeauthoryear{Zhao, Zhang, and Wang}{Zhao et~al\mbox{.}}{2021}]%
        {zhao2021graphsmote}
\bibfield{author}{\bibinfo{person}{Tianxiang Zhao}, \bibinfo{person}{Xiang Zhang}, {and} \bibinfo{person}{Suhang Wang}.} \bibinfo{year}{2021}\natexlab{}.
\newblock \showarticletitle{Graphsmote: Imbalanced node classification on graphs with graph neural networks}. In \bibinfo{booktitle}{\emph{Proceedings of the 14th ACM international conference on web search and data mining}}. \bibinfo{pages}{833--841}.
\newblock


\bibitem[\protect\citeauthoryear{Zhao, Zhou, Li, Tang, Wang, Hou, Min, Zhang, Zhang, Dong, et~al\mbox{.}}{Zhao et~al\mbox{.}}{2023a}]%
        {zhao2023survey}
\bibfield{author}{\bibinfo{person}{Wayne~Xin Zhao}, \bibinfo{person}{Kun Zhou}, \bibinfo{person}{Junyi Li}, \bibinfo{person}{Tianyi Tang}, \bibinfo{person}{Xiaolei Wang}, \bibinfo{person}{Yupeng Hou}, \bibinfo{person}{Yingqian Min}, \bibinfo{person}{Beichen Zhang}, \bibinfo{person}{Junjie Zhang}, \bibinfo{person}{Zican Dong}, {et~al\mbox{.}}} \bibinfo{year}{2023}\natexlab{a}.
\newblock \showarticletitle{A survey of large language models}.
\newblock \bibinfo{journal}{\emph{arXiv preprint arXiv:2303.18223}} (\bibinfo{year}{2023}).
\newblock


\bibitem[\protect\citeauthoryear{Zheng, Chiang, Sheng, Zhuang, Wu, Zhuang, Lin, Li, Li, Xing, et~al\mbox{.}}{Zheng et~al\mbox{.}}{2023}]%
        {zheng2023judging}
\bibfield{author}{\bibinfo{person}{Lianmin Zheng}, \bibinfo{person}{Wei-Lin Chiang}, \bibinfo{person}{Ying Sheng}, \bibinfo{person}{Siyuan Zhuang}, \bibinfo{person}{Zhanghao Wu}, \bibinfo{person}{Yonghao Zhuang}, \bibinfo{person}{Zi Lin}, \bibinfo{person}{Zhuohan Li}, \bibinfo{person}{Dacheng Li}, \bibinfo{person}{Eric Xing}, {et~al\mbox{.}}} \bibinfo{year}{2023}\natexlab{}.
\newblock \showarticletitle{Judging LLM-as-a-judge with MT-Bench and Chatbot Arena}.
\newblock \bibinfo{journal}{\emph{arXiv preprint arXiv:2306.05685}} (\bibinfo{year}{2023}).
\newblock


\bibitem[\protect\citeauthoryear{Zhong, Liu, Ao, Hu, Feng, Tang, and He}{Zhong et~al\mbox{.}}{2020}]%
        {zhong2020financial}
\bibfield{author}{\bibinfo{person}{Qiwei Zhong}, \bibinfo{person}{Yang Liu}, \bibinfo{person}{Xiang Ao}, \bibinfo{person}{Binbin Hu}, \bibinfo{person}{Jinghua Feng}, \bibinfo{person}{Jiayu Tang}, {and} \bibinfo{person}{Qing He}.} \bibinfo{year}{2020}\natexlab{}.
\newblock \showarticletitle{Financial defaulter detection on online credit payment via multi-view attributed heterogeneous information network}. In \bibinfo{booktitle}{\emph{Proceedings of The Web Conference 2020}}. \bibinfo{pages}{785--795}.
\newblock


\bibitem[\protect\citeauthoryear{Zhou, Huang, Song, Chen, and Hu}{Zhou et~al\mbox{.}}{2022}]%
        {zhou2022auto}
\bibfield{author}{\bibinfo{person}{Kaixiong Zhou}, \bibinfo{person}{Xiao Huang}, \bibinfo{person}{Qingquan Song}, \bibinfo{person}{Rui Chen}, {and} \bibinfo{person}{Xia Hu}.} \bibinfo{year}{2022}\natexlab{}.
\newblock \showarticletitle{Auto-gnn: Neural architecture search of graph neural networks}.
\newblock \bibinfo{journal}{\emph{Frontiers in big Data}}  \bibinfo{volume}{5} (\bibinfo{year}{2022}), \bibinfo{pages}{1029307}.
\newblock


\bibitem[\protect\citeauthoryear{Zhu, Yan, Zhao, Heimann, Akoglu, and Koutra}{Zhu et~al\mbox{.}}{2020}]%
        {zhu2020beyond}
\bibfield{author}{\bibinfo{person}{Jiong Zhu}, \bibinfo{person}{Yujun Yan}, \bibinfo{person}{Lingxiao Zhao}, \bibinfo{person}{Mark Heimann}, \bibinfo{person}{Leman Akoglu}, {and} \bibinfo{person}{Danai Koutra}.} \bibinfo{year}{2020}\natexlab{}.
\newblock \showarticletitle{Beyond homophily in graph neural networks: Current limitations and effective designs}.
\newblock \bibinfo{journal}{\emph{Advances in neural information processing systems}}  \bibinfo{volume}{33} (\bibinfo{year}{2020}), \bibinfo{pages}{7793--7804}.
\newblock


\bibitem[\protect\citeauthoryear{Zhu, Wang, Shi, Ji, and Cui}{Zhu et~al\mbox{.}}{2021}]%
        {zhu2021interpreting}
\bibfield{author}{\bibinfo{person}{Meiqi Zhu}, \bibinfo{person}{Xiao Wang}, \bibinfo{person}{Chuan Shi}, \bibinfo{person}{Houye Ji}, {and} \bibinfo{person}{Peng Cui}.} \bibinfo{year}{2021}\natexlab{}.
\newblock \showarticletitle{Interpreting and unifying graph neural networks with an optimization framework}. In \bibinfo{booktitle}{\emph{Proceedings of the Web Conference 2021}}. \bibinfo{pages}{1215--1226}.
\newblock


\bibitem[\protect\citeauthoryear{Zhu, Wu, Guo, Hong, and Li}{Zhu et~al\mbox{.}}{2024}]%
        {zhu2024collaborative}
\bibfield{author}{\bibinfo{person}{Yaochen Zhu}, \bibinfo{person}{Liang Wu}, \bibinfo{person}{Qi Guo}, \bibinfo{person}{Liangjie Hong}, {and} \bibinfo{person}{Jundong Li}.} \bibinfo{year}{2024}\natexlab{}.
\newblock \showarticletitle{Collaborative large language model for recommender systems}. In \bibinfo{booktitle}{\emph{Proceedings of the ACM on Web Conference 2024}}. \bibinfo{pages}{3162--3172}.
\newblock


\end{thebibliography}

\end{document}